\documentclass{article}

\usepackage{booktabs} 

\usepackage{hyperref}

\usepackage{color, colortbl}
\usepackage[accepted]{icml2024}

\usepackage{amsmath,amsfonts,bm}

\def\eqref#1{equation~\ref{#1}}

\def\1{\bm{1}}

\DeclareMathAlphabet{\mathsfit}{\encodingdefault}{\sfdefault}{m}{sl}
\SetMathAlphabet{\mathsfit}{bold}{\encodingdefault}{\sfdefault}{bx}{n}

\usepackage{url}            
\usepackage{amsfonts}       
\usepackage{nicefrac}       
\usepackage{microtype}      
\usepackage{graphicx}
\usepackage{float}
\usepackage{subfig}
\usepackage{multirow}
\usepackage{enumitem}
\usepackage{amsmath}
\usepackage{amssymb}
\usepackage{mathtools}
\usepackage{amsthm}
\usepackage[noend]{algpseudocode}

\renewcommand{\algorithmicrequire}{\textbf{Input:}}
\renewcommand{\algorithmicensure}{\textbf{Output:}}

\newcommand{\methodname}{BA$^2$-Det}

\definecolor{dgreen}{RGB}{1,150,74}
\newcommand\up[1]{\textcolor{dgreen}{$\uparrow{#1}$}}
\newcommand\down[1]{\textcolor{red}{$\downarrow{#1}$}}
\definecolor{darkgreen}{HTML}{009B55}
\definecolor{linkcolor}{HTML}{ED1C24}
\definecolor{bluegreen}{HTML}{00B3B8}

\usepackage[capitalize,noabbrev]{cleveref}

\theoremstyle{plain}

\theoremstyle{definition}

\theoremstyle{remark}

\icmltitlerunning{Weakly Supervised 3D Object Detection with Multi-Stage Generalization}

\begin{document}

\twocolumn[
\icmltitle{	
Weakly Supervised 3D Object Detection with Multi-Stage Generalization}




\begin{icmlauthorlist}
\icmlauthor{Jiawei He}{1,2}
\icmlauthor{Yuqi Wang}{1,2}
\icmlauthor{Yuntao Chen}{3}
\icmlauthor{Zhaoxiang Zhang}{1,2,3}
\end{icmlauthorlist}

\icmlaffiliation{1}{CRIPAC, Institute of Automation, Chinese Academy of Sciences (CASIA)}
\icmlaffiliation{2}{School of Artificial Intelligence, University of Chinese Academy of Sciences (UCAS)}
\icmlaffiliation{3}{Centre for Artificial Intelligence and Robotics, HKISI\_CAS}

\icmlcorrespondingauthor{Jiawei He}{hejiawei2019@ac.cn}
\icmlcorrespondingauthor{Yuqi Wang}{wangyuqi2020@ac.cn}
\icmlcorrespondingauthor{Yuntao Chen}{chenyuntao08@gmail.com}
\icmlcorrespondingauthor{Zhaoxiang Zhang}{zhaoxiang.zhang@ia.ac.cn}

\icmlkeywords{Machine Learning, ICML}

\vskip 0.3in
]



\printAffiliationsAndNotice{}  

\begin{abstract}
With the rapid development of large models, the need for data has become increasingly crucial. Especially in 3D object detection, costly manual annotations have hindered further advancements.
To reduce the burden of annotation, we study the problem of achieving 3D object detection solely based on 2D annotations.
Thanks to advanced 3D reconstruction techniques, it is now feasible to reconstruct the overall static 3D scene. However, extracting precise object-level annotations from the entire scene and generalizing these limited annotations to the entire scene remain challenges.
In this paper, we introduce a novel paradigm called \methodname{}, encompassing pseudo label generation and multi-stage generalization. 
We devise the DoubleClustering algorithm to obtain object clusters from reconstructed scene-level points, and further enhance the model's detection capabilities by developing three stages of generalization: progressing from complete to partial, static to dynamic, and close to distant.
Experiments conducted on the large-scale Waymo Open Dataset show that the performance of \methodname{} is on par with the fully-supervised methods using 10\% annotations. 
Additionally, using large raw videos for pretraining, \methodname{} can achieve a 20\% relative improvement on the KITTI dataset. The method also has great potential for detecting open-set 3D objects in complex scenes. Project page: \url{https://ba2det.site}.
\end{abstract}

\section{Introduction}
3D object detection has gained increasing attention from researchers and has become a fundamental task in real-world perception.
Thanks to the efforts of researchers, fully supervised 3D object detection is practical in both traffic scenes~\citep{shi2020pv, yin2021center, fan2023super} and indoor scenes~\citep{qi2019deep,liu2021group}.
In recent years, the cheap cost of camera sensors has fostered the emergence of image-based 3D object detection as a rapidly developing field of research, including monocular and multi-camera settings.
In terms of performance, camera-only 3D object detectors~\citep{Wang_2023_ICCV} are also gradually catching up with LiDAR-based methods.\par
However, due to the requirement of extensive and costly manual annotations, the further development of fully supervised 3D object detection methods is potentially limited. 
To overcome this limitation, some previous works have explored weak supervision algorithms~\citep{zakharov2020autolabeling,peng2022weakm3d} with additional LiDAR data to unlock the potential of unlabeled images.
However, the reliance on LiDAR sensors limits the practicality of these methods in more general scenarios. With the advancement of 2D foundation models~\citep{kirillov2023segment}, 2D annotation is no longer a bottleneck.
In this paper, we aim to investigate the feasibility of achieving 3D object detection solely based on 2D annotations, an unexplored problem.\par

The core challenge for this problem lies in deriving 3D information from 2D images. Drawing inspiration from 3D reconstruction techniques~\citep{schonberger2016structure}, we can obtain the overall static 3D scene structure.
Therefore, the core challenge has shifted to \emph{extracting object-level pseudo-labels from the global scene}, and \emph{generalizing limited object pseudo-labels to more objects}.\par

To address this challenge, we introduce a novel paradigm called \methodname{}, consisting of Pseudo Label Generation and Multi-Stage Generalization. 
The Pseudo Label Generation phase generates object-level 3D bounding boxes based on the reconstructed 3D scene and 2D bounding boxes in each frame. We have devised a two-step clustering algorithm called \emph{DoubleClustering}, consisting of intra-frame Local Point Clustering (LPC) and inter-frame Global Point Clustering (GPC). This algorithm ensures more complete and clean object clusters, from which we can easily obtain initial object pseudo-labels.
However, these initial object pseudo-labels are only accurate for static and fully reconstructed objects. 
This is because moving objects cannot be reconstructed in scene-level 3D reconstruction, and some object clusters may be incomplete due to occlusion or far distance. 
To tackle this issue, our Multi-Stage Generalization leverages the generalization capability of neural network models. 
The first stage of generalization is from complete objects to partial ones. We learn the complete 3D bounding box of partial objects from other non-occluded 3D boxes using a PointNet-like neural network. 
The second stage of generalization is from static to dynamic. We develop a monocular 3D object detector that operates on single-frame images. This detector is capable of learning movement-agnostic 3D representations from monocular images, as there is no visual distinction between moving and stationary objects in an image. Additionally, we have devised a new label assignment strategy, orientation loss, and an iterative self-retraining strategy for efficient training of the detector using pseudo-labels.
Finally, we also employ temporal aggregation, inspired by \cite{he2022badet}, to enhance close-to-distant generalization.
With the help of these generalizations, our method can effectively adapt to full-scene 3D object detection.
In summary, our main contributions are as follows: 
\begin{itemize}
    \item We propose a novel paradigm for weakly supervised monocular 3D object detection using only 2D labels. Leveraging 3D reconstruction and the generalization capabilities of neural networks, we present a practical solution to this problem for the first time.
    \item Our proposed method, named \methodname{}, addresses three fundamental technical challenges in learning the 3D object detector. We have developed three stages of generalizations: from complete to partial, from static to dynamic, and from close to distant.
    \item We conducted experiments on various datasets, including the KITTI dataset and the large-scale Waymo Open Dataset (WOD), demonstrating the effectiveness of our method for generating high-quality 3D labels and leveraging large-scale data for pretraining. The performance of \methodname{} is comparable to the fully-supervised BA-Det trained with only 10\% of the videos, and even outperforms some leading fully-supervised methods. As a pretraining approach, \methodname{} can achieve a relative improvement of 20\% on the KITTI dataset.
    \item We further investigate the potential impact of our method, including the detection of open-set 3D objects in complex scenes and the downstream application of 3D object tracking.
\end{itemize}
\section{Related Work}
\subsection{Fully Supervised Monocular 3D Object Detection}
Monocular 3D object detection~\citep{chen2016monocular, brazil2019m3d, wang2019pseudo, zhang2021objects} has been explored for several years.
The existing monocular 3D object detection methods can be divided into three categories: regressing 3D objects from the image directly, regressing on the depth map or lifted 3D space, and regressing based on geometric constraints. CenterNet~\citep{zhou2019objects} and FCOS3D~\citep{wang2021fcos3d} are the representing works to estimate the 3D objects with a 3D regression branch based on the 2D object detectors. 
PL~\citep{wang2019pseudo} and PL++~\citep{You2020Pseudo-LiDAR++:} use the off-the-shelf dense depth estimator to project the scene in 3D space and detect objects from pseudo-LiDAR.
D4LCN~\citep{ding2020learning} and PatchNet~\citep{ma2020rethinking} use the image-aligned depth map to extract features.
E2E-PL~\citep{qian2020end} and CaDDN~\citep{reading2021categorical} jointly learn the depth estimator and 3D object detector in an end-to-end manner.
With the geometric constraints, the 3D object depth can be estimated by solving Perspective-n-Point (PnP) problem. 
MonoFlex~\citep{zhang2021objects} solves PnP from the vertical lines of 3D bounding boxes. 
DCD~\citep{li2022densely} uses arbitrary point pairs to construct dense constraints.
Recently, there has been a surge in the development of temporal 3D object detection from monocular images.
DfM~\citep{wang2022monocular} and BA-Det~\citep{he2022badet} aggregate temporal information at the scene level and in an object-centric manner, inspired by two-view and multi-view geometry theory.\par
\begin{figure*}[t]
          \begin{center}
           \includegraphics[width=\linewidth]{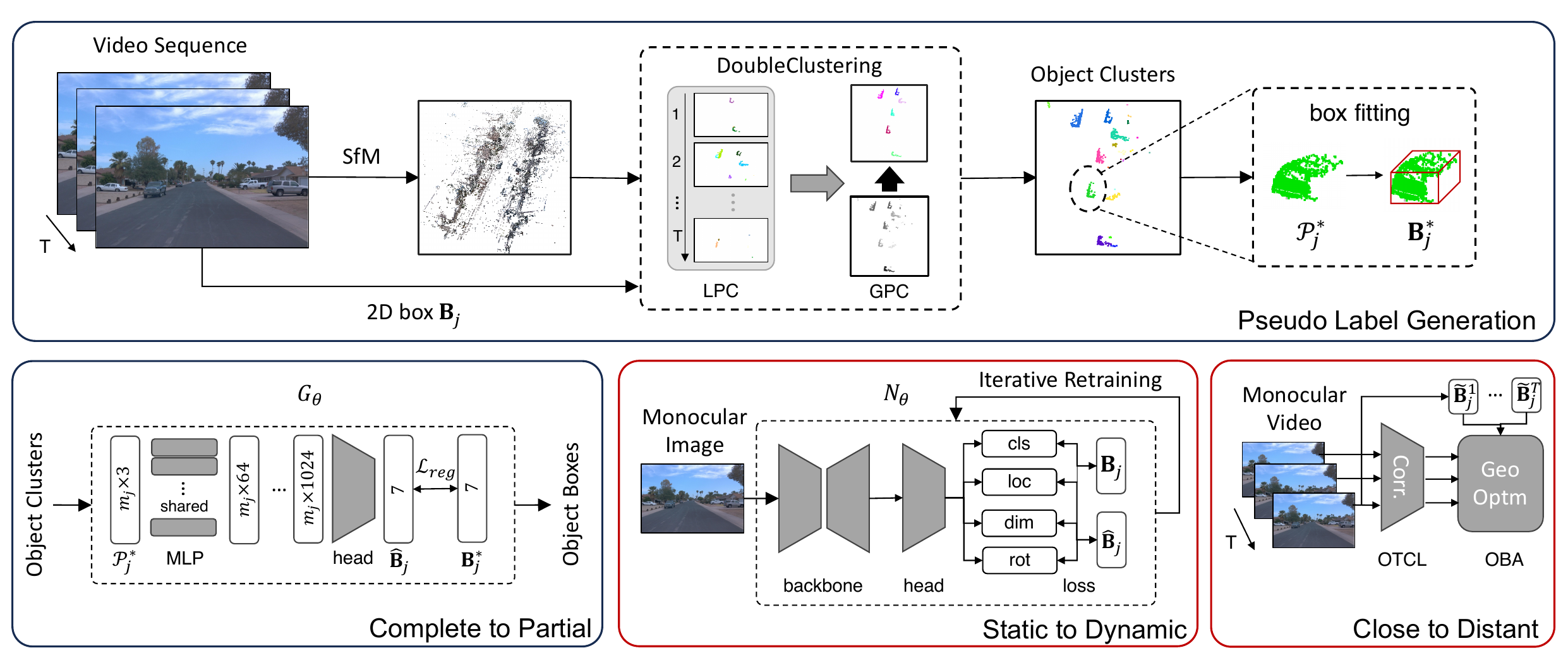}
           \end{center}
           \vspace{-8pt}
             \caption{\textbf{Pipeline of \methodname{}.} Top: reconstruction-based pseudo label generation process. We cluster the object point clouds from the reconstructed scene and fit the tight bounding box as the pseudo label. Bottom: Three stages of network generalization. The neural networks inside the \textcolor[rgb]{0.75,0,0}{red rounded rectangles} are also for the inference.}
             \label{fig:pipeline}
             \vspace{-10pt}
\end{figure*}  
\subsection{Detecting 3D Objects without 3D Labels}
Since the 3D object detection task made great progress in the past few years, many researchers have begun to explore using fewer 3D labels or even without 3D labels to train a 3D object detector.
For LiDAR-based 3D object segmentation, clustering-based methods~\citep{triebel2010segmentation,campello2013density,nunes2022unsupervised} are the mainstream methods. LSMOL~\citep{wang20224d} and \citet{najibi2022motion} combine image and LiDAR to segment 2D and 3D objects. However, only class-agnostic segmentation can be achieved in these methods.
MODEST~\citep{you2022learning} is an unsupervised 3D mobile object detection method to predict 3D bounding box. Its key idea is that mobile objects are ephemeral members of a scene.
For image-based 3D object detection, SDFLabel~\citep{zakharov2020autolabeling} is a pioneer work that can auto-label the 3D bounding boxes from a pre-trained 2D detector and the corresponding LiDAR data by recovering the object shape with signed distance fields (SDF).
WeakM3D~\citep{peng2022weakm3d} is also a weakly supervised method and needs additional LiDAR data. \citet{yang2022towards} first explore the image-only weakly supervision without LiDAR. However, box size and orientation cannot be estimated in this method and can only be learned by ground truth in the semi-supervised setting.
Unlike the above works, our \methodname{} \emph{only uses images} without LiDAR as an auxiliary modality and can estimate 3D \emph{bounding boxes} including center position, box size, and orientation.\par
Besides, there are some weakly supervised methods~\citep{yang2021dsc,li2022ws} designed for the 6DoF pose estimation task. However, these two tasks encounter distinct challenges. In 6DoF pose estimation, the models learn the rotation and translation of specific objects on the table or near the camera. The primary challenge in 3D object detection is accurately estimating the depth and pose of objects in distant and heavily occluded scenes. So there are significant differences in methodology and experimental details between 6DoF pose estimation and 3D object detection.
Moreover, in our setting, we do not require additional 3D models (class-aware CAD models) as prior knowledge, which is the common practice in 6DoF pose estimation. 
\section{Methodology}
\paragraph{Problem setup.} In this paper, we present the task of weakly supervised monocular 3D object detection with only 2D labels. Our method is designed not to require detailed 2D object masks for supervision; only 2D bounding boxes are sufficient. The object predictions are represented as 7-dimensional 3D bounding boxes, i.e., object center location $(c_x,c_y,c_z)$ in the camera frame, 3D box size $(w,h,l)$, and yaw rotation $r_y$. The main difference between previous studies and ours is that our detector does not require training with 3D ground truth or the use of LiDAR data as an auxiliary modality. Our \methodname{} also does not rely on manually crafted CAD models and pretrained depth models. 

\paragraph{Algorithm overview.}
We introduce our framework \methodname{} briefly and explain module designs. As shown in Fig.~\ref{fig:pipeline}, \methodname{} includes two main parts, high-quality pseudo label generation and multi-stage generalization from the pseudo labels. In \textbf{pseudo label generation} phase (Sec.~\ref{sec:doublecluster}), we first utilize scene-level reconstruction from the moving camera to obtain the global point clouds. To extract 3D object clusters from scene reconstruction, we design a \textit{DoubleClustering} algorithm. The object clusters are further fitted with cuboids to form 3D bounding boxes. During the subsequent \textbf{multi-stage generalization} phase (Sec.~\ref{sec:global}), (1) To generalize \textit{from complete to partial}, we develop a neural network to learn 3D object bounding boxes for partial objects from well-reconstructed objects; (2) To generalize \textit{from static to dynamic}, we train a 3D object detector with carefully designed learning strategies and iterative refinement; (3) To generalize \textit{from close to distant}, we follow learn temporal object detector with geometric feature aggregation.\par
\subsection{Pseudo Label Generation from Scene-level Reconstruction}
\label{sec:doublecluster}
Using the Structure-from-Motion (SfM) technique, it is possible to reconstruct a 3D scene from ego-motion.
Then from the reconstructed scene, with the help of 2D bounding boxes in each frame, the 3D object cluster can be obtained by clustering the foreground points from the reconstructed scene.
So, in this section, we introduce an algorithm called DoubleClustering (Alg.~\ref{alg:cluster}) for extracting 3D object clusters from the 3D reconstructed scene. Then we optimize and generate tight 3D bounding boxes for each object cluster as an initial pseudo label.\par
\paragraph{Scene Reconstruction.}Firstly, let's revisit scene reconstruction with SfM.
 We denote the video sequence as $\mathcal{V}=\{\mathbf{I}_t|t=1,2\cdots,T\}$, keypoints in image $\mathbf{I}_t$ as $\mathbf{p}_{t}^i=[u_i,v_i]^\top,(i=1,2,\cdots,n)$ and local feature on each keypoint as $\mathcal{F}_t=\{\mathbf{f}_{t}^i\}$. In this paper, we use the keypoint extractor and local feature-matching network SuperPoint~\citep{detone2018superpoint} and SuperGlue~\citep{sarlin2020superglue}. 
Given intrinsic parameter $\mathbf{K}$ and the extrinsic parameter $\mathbf{T}_{t}=[\mathbf{R}_{t}|\mathbf{t}_{t}]$ of the camera in time $t$, 3D keypoint $\mathbf{P}_{i}$ in the global frame can be optimized by solving bundle adjustment (BA) with projection error computed on the corresponding keypoints between any two images as 
\begin{equation}
    \{{\mathbf{P}}^*_{i}\}_{i=1}^n=\underset{\{\mathbf{P}_i\}_{i=1}^n}{\arg\min}\frac{1}{2}\sum_{i=1}^{n}\sum_{t=1}^{T}||\mathbf{p}_i^t-\Pi(\mathbf{T}_t,\mathbf{P}_i,\mathbf{K})||^2,
\end{equation}
where $\Pi(\cdot)$ is the function projecting the 3D points in the world frame to the image. Besides, $\mathbf{K}$ and $\mathbf{T}_{t}$ can be also optimized in BA process.
Please note that when the ego moves slowly, the disparity between two frames is small and the observation noise can affect reconstruction. Therefore, when the camera moves slowly, we disregard the video sequence and do not reconstruct the scene. The speed threshold is defined as $\omega$.\par
\begin{algorithm}[t]
	\caption{Generating object clusters (DoubleClustering) }
	\label{alg:cluster}
 
  \begin{algorithmic}[1]
	\Require
         video clip $\mathcal{V}$, camera intrinsic $\{\mathbf{K}\}$, camera pose $\{\mathbf{T}_{t}\}$, 2D bounding box $\{\mathbf{B}_j^t\}_{j=1}^{n_d}$
         \Ensure
	3D object cluster $\{\mathcal{P}_{j}^*\}_{j=1}^{n_d}$
        \State $\{{\mathbf{P}}^*_{i}\}_{i=1}^n \leftarrow \mathtt{SfM}(\mathcal{V},\{\mathbf{K}\},\{\mathbf{T}_{t}\})$;
        \State $\mathcal{P}\leftarrow \varnothing$;
	\For{$j \in [1,n_d]$} \Comment{LPC for each object in each frame}
        \For{$t \in [1,T]$}
        \State $\mathcal{P}_j^t\leftarrow \mathtt{LPC}(\{{\mathbf{P}}^*_{i}\},\mathbf{B}_j^t)$;
        \State $\mathcal{P}\leftarrow \mathcal{P}\cup \mathcal{P}_j^t$;
        \EndFor
        \EndFor
        
        \State $\{\mathcal{P}_{j'}\}_{j'=1}^{n'_d}\leftarrow \mathtt{GPC}(\mathcal{P})$;
        \Comment{GPC from $\mathcal{P}$ to generate $n_d'$ clusters}
        \State $\{\mathcal{P}_{j}^*\}_{j=1}^{n_d}\leftarrow \mathtt{Match}(\{\mathcal{P}_{j'}\},\{\mathbf{B}_j^t\})$;
        \Comment{Match objects to the point clusters}
	\State \Return {$\{\mathcal{P}_{j}^*\}_{j=1}^{n_d}$}
       \end{algorithmic} 
\end{algorithm}
\paragraph{Object point cloud clustering.}After the scene reconstruction, we introduce a two-step object clustering algorithm, called DoubleClustering, to separate and cluster the object point clouds in the reconstructed scene.
Firstly, in each frame, we choose the 3D points that can be projected in the 2D bounding boxes, and perform the Local Point Clustering (LPC) to choose the largest cluster for each object $\mathbf{B}_j^t$:
\begin{equation}
    \mathcal{P}_j^t=\{{\mathbf{P}}_{i}\}_{i=1}^m = \mathtt{LPC}(\{\mathbf{P}_i^*|\Pi(\mathbf{T}_t,\mathbf{P}_i^*,\mathbf{K})\in \mathbf{b}_j^t\}),
\end{equation}
where we denote $\mathbf{b}_j^t$ as the image region in the 2D box $\mathbf{B}_j^t$. The cluster algorithm is based on Connected Component (CC) algorithm, and the distance threshold in CC algorithm is $\delta_1$.\par
Secondly, we gather the clusters from each frame, and conduct global clustering in the whole scene.
The clustering algorithm is also based on CC with distance threshold $\delta_2$, called Global Point Clustering (GPC):
\begin{equation}
    \{\mathcal{P}_{j'}\}_{j'=1}^{n'_d} = \mathtt{GPC}(\bigcup_{t,j}\mathcal{P}_j^t),
\end{equation}
where $n_d'$ is the total object clusters. We ignore clusters with point numbers lower than threshold $\theta$, as they may represent noise points.
Finally, we choose the object cluster with the highest number of projected points in it as the corresponding cluster $\mathcal{P}_{j}^*$ for the 2D bounding box $\mathbf{B}_j^t$.

\paragraph{3D bounding box fitting.}
\label{sec:fit}
We now obtain object clusters and then we need to generate the initial 3D pseudo box from the object points in each cluster. 
Given the object cluster $\mathcal{P}_{j}^*$, we fit a tight 3D bounding box according to the object points. These tight boxes serve as the initial 3D labels.
Utilizing the assumption that the reconstructed points are mainly located on the object's surface and taking inspiration from~\citet{zhang2017efficient}, we optimize the orientation $r_y$ by minimizing the total distance between the points and their closest edge. Subsequently, we adjust the width and length of the bird's-eye view bounding box to achieve minimal area:
\begin{equation}
    r_y^*,\mathbf{B}_{bev}^* = \mathop{\arg\min}_{r_y\in[0,\pi),\mathbf{B}_{bev}\in\mathbb{R}^2}\sum_{i=1}^m\min_{l\in[1,4]}d(\mathbf{P}_i,\mathbf{R}(r_y)\mathbf{B}_{bev}^l),
\end{equation}
where $\mathbf{P}_i$ is the 3D points in the object cluster, and $\mathbf{R}(r_y)\mathbf{B}_{bev}^l$ is the edge of rotated bounding box in Bird's Eye View (BEV) with angle $r_y$, we use $l_2$ distance as distance function $d(\cdot)$. We calculate the box's height using points along the z-axis.We calculate the box height by measuring the distance between the highest and lowest points of the point cloud along the z-axis, and finally generate the 3D box $\mathbf{B}_j^*$ for object $j$.\par
\subsection{Multi-Stage Generalization}
\label{sec:global}
Some objects may be occluded or influenced by outlier points. These partial objects have not been reconstructed well, resulting in inaccurate labels, especially for size and orientation estimation. So we design the first-stage complete-to-partial generalization.
 In the traditional SfM system, reconstructing points on static objects is easy, but it is challenging to reconstruct moving objects due to their different movement from ego-motion and being filtered by epipolar geometry constraints. Only \textit{static} objects are in initial pseudo labels. So we propose the second-stage static-to-dynamic generalization.
 Besides, distant objects have fragile visual features. So we propose the third-stage close-to-distant temporal generalization.\par
\begin{figure}[t]
         \begin{center}
          \resizebox{\linewidth}{!}{
	\includegraphics[width =\linewidth]{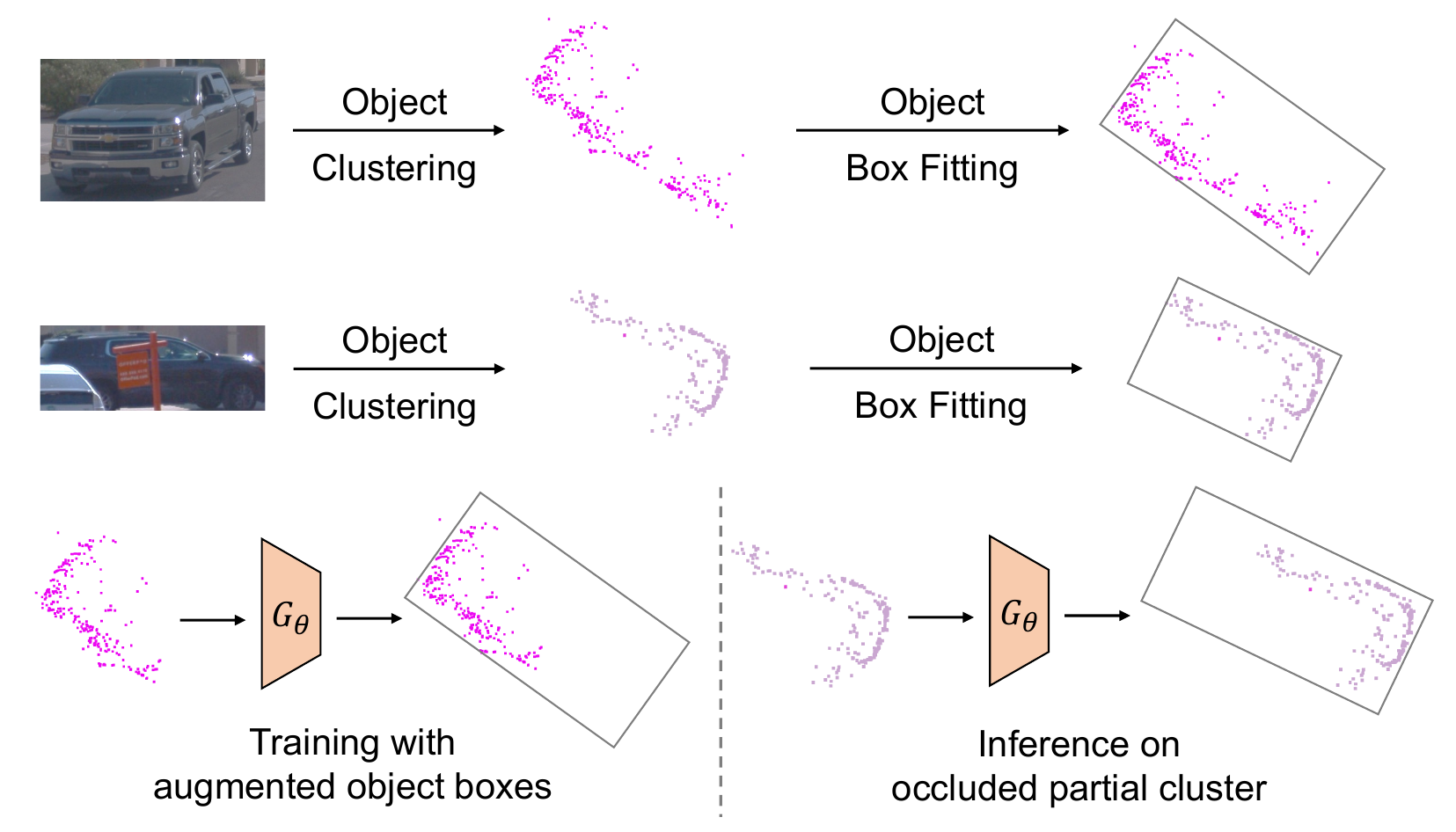}
         }
        \end{center}
           \vspace{-10pt}
             \caption{\textbf{Illustrations of fitted bounding boxes and generalized boxes.} Some occluded objects are badly reconstructed, leading to inaccurate pseudo-labels. $G_\theta$ can generalize from augmented complete objects to partial objects. }
             \label{fig:g_theta}
             \vspace{-10pt}
\end{figure}

 \paragraph{Generalization from complete to partial objects.}
 \label{sec:g_theta}
 The first stage of generalization is to train a model to predict a full 3D bounding box from occluded/partial object point clouds.
 Therefore, we design a neural network $G_\theta$ to learn and refine the initial 3D bounding box from well-reconstructed complete objects. 
 In general, we expect to first find the well-reconstructed/complete objects as training data, i.e., the objects whose 3D bounding box can be correctly fitted from the object point clouds. Then learning from these objects, the network can predict full 3D boxes for all object clusters. Fig~\ref{fig:g_theta} shows how does $G_\theta$ work.\par
 This network takes the object cluster $\mathcal{P}_{j}^*\in \mathbb{R}^{m_j\times 3}$ as input and normalizes the coordinates of 3D points using the center of the initial 3D pseudo box $\mathbf{B}_j^*$.
It consists of a PointNet backbone and a head to predict 7DoF 3D bounding box $\widehat{\mathbf{B}}_j=[c_x,c_y,c_z,w,h,l,r_y]$. We use the smooth L1 bounding box localization loss $\mathcal{L}_{reg}$. Note that we only consider the length between $[\sigma_0,\sigma_1]$ as the well-reconstructed object and take these 3D pseudo boxes to supervise $G_\theta$.
To simulate the occlusion-induced partial object points, we perform data augmentation by randomly cutting off regions from the well-reconstructed object cluster.  More details in training $G_\theta$ are in Sec.~\ref{sec:gba_detail} in the appendix.\par
\paragraph{Generalization from static to moving objects.}
 \begin{figure}[t]
         \begin{center}
          \resizebox{0.9\linewidth}{!}{
	\includegraphics[width =\linewidth]{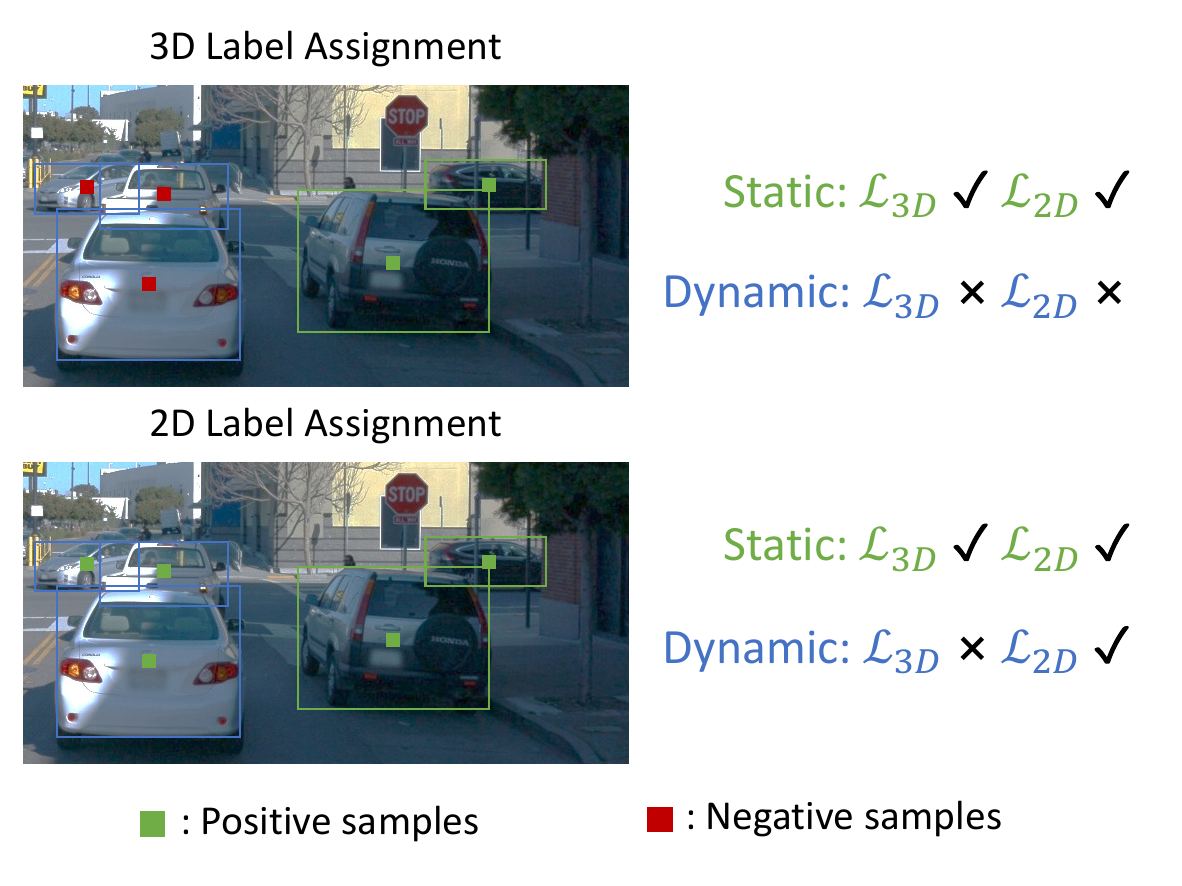}
         }
        \end{center}
           \vspace{-15pt}
             \caption{\textbf{Comparison between 2D and 3D label assignment.} For 3D label assignment used in fully supervised setting, moving objects are negative samples. However, our 2D label assignment keeps them as positive samples.}
             \label{fig:label_assign}
             \vspace{-10pt}
\end{figure} 
We still need to address the issue that pseudo labels mainly come from static objects.
We observe that in a single monocular image, static and moving objects have similar appearances.
Therefore, the network can generalize the 3D object predictions learning from static labels to other moving objects.
We denote the monocular 3D object detector as $N_\theta$. The network architecture is based on CenterNet~\citep{zhou2019objects,zhang2021objects}. 
In our weakly supervised setting, many objects have 2D labels ${\mathbf{B}}_j$, but there are no corresponding 3D pseudo labels $\widehat{\mathbf{B}}_j$ due to their movement. Besides, the pseudo labels $\widehat{\mathbf{B}}_j$ may have inaccurate orientation.
So, different from a fully supervised object detector, we design a new label assignment strategy and orientation loss. \par
Regarding the issue of unlabeled moving objects, training the network with traditional 3D label assignment will lead to many false negatives.
We assign labels using 2D ground truth (GT) labels and disregard their 3D losses if there are no 3D pseudo labels as
\begin{equation}
    \mathcal{L}_{3D}(u,v)= 
    \left\{\begin{aligned} 
    &0, &&\text{if}\ (u,v)\notin\mathcal{C}, \\
&\mathcal{L}_{loc}+\mathcal{L}_{dim}+\mathcal{L}_{r},&&\text{else},
\end{aligned}\right.
\label{eq:3dass}
\end{equation}
which $(u,v)$ is the pixel on the image, and $\mathcal{C}=\{(c_u,c_v)| [(c_u,c_v)\ \text{is center of}\ \mathbf{B}_j ]\wedge [\widehat{\mathbf{B}}_j\neq\varnothing ]\}$.
As shown in Fig.~\ref{fig:label_assign}, the proposed 2D assignment ensures that the unlabeled objects are not considered negative samples. \par
Another difficulty for weak supervision is to distinguish whether the object is facing forward or backward. The orientation of the 3D pseudo label may have a deviation of $180^{\circ}$ from the real heading.
We modify the original MultiBin orientation loss~\citep{mousavian20173d} to alleviate this problem. The new orientation loss is the minimum in original loss and $180^{\circ}$-reversed one: 
\begin{equation}
    \mathcal{L}_{r}= \min(\mathcal{L}_{M}(\hat{r}_y,r_y),\mathcal{L}_{M}(\pi+\hat{r}_y,r_y)),
\end{equation}
where $\mathcal{L}_{M}$ is MultiBin Loss, $\hat{r}_y$ is yaw prediction, $r_y$ is pseudo ground truth.\par
\begin{table*}[]
\centering
\resizebox{\linewidth}{!}{
\begin{tabular}{l|c|c|cccc|ccc}
\specialrule{1pt}{0pt}{1pt}
\toprule
Method&Tem.&3D Sup.& 3D AP$_\text{5}$ & 3D APH$_\text{5}$ & 3D AP$_\text{50}$ & 3D APH$_\text{50}$ & LET APL$_\text{50}$ & LET AP$_\text{50}$ & LET APH$_\text{50}$\\
\midrule
PatchNet~\citep{ma2020rethinking}&&100\%$^\dag$&- &- & 2.92&2.74 &-&-&-\\
M3D-RPN~\citep{brazil2019m3d} &&100\%$^\dag$ &- &- &3.79 & 3.63 &-&-&-\\
PCT~\citep{pct}&&100\%$^\dag$&- &-  &4.20 &4.15 &-&-&-\\
MonoJSG~\citep{lian2022monojsg}&&100\%$^\dag$&- &-  &5.65 &\textbf{5.47} &-&-&-\\
\textbf{\methodname{} (Ours)} &&0\%$^\dag$ & \textbf{55.24} & \textbf{40.87} & \textbf{6.24} & 5.37 & \textbf{16.61} & \textbf{27.94} & \textbf{21.32} \\
\midrule
\textcolor{gray}{MonoFlex~\citep{zhang2021objects}}& &\textcolor{gray}{100\%}&\textcolor{gray}{70.33} &\textcolor{gray}{69.41} &\textcolor{gray}{34.70} &\textcolor{gray}{34.43} &\textcolor{gray}{50.63} &\textcolor{gray}{67.30} &\textcolor{gray}{66.50}\\
\textcolor{gray}{BA-Det~\citep{he2022badet}}&\textcolor{gray}{\checkmark} &\textcolor{gray}{100\%}&\textcolor{gray}{72.96} &\textcolor{gray}{71.78} &\textcolor{gray}{40.93} &\textcolor{gray}{40.51} &\textcolor{gray}{54.45} &\textcolor{gray}{68.32} &\textcolor{gray}{67.36}\\
MonoFlex~\citep{zhang2021objects}& &10\% &53.68 &52.30 &15.44 &15.22 &28.21 &44.21&43.23\\
BA-Det~\citep{he2022badet}&\checkmark&10\% &57.29 &55.27&19.70 &19.27 &32.53 &46.91 &45.52\\
\textbf{\methodname{} (Ours)}&\checkmark &10\%&\textbf{65.24}	&\textbf{62.95}&	\textbf{25.17}	&\textbf{24.65}	&\textbf{42.55}&	\textbf{58.77}	&\textbf{57.14}\\
\midrule
SfM+BA-Det (Baseline)&\checkmark&0\% &27.84 &8.80 &2.89 &0.75 &7.34 &10.75 &3.31\\
\textbf{\methodname{} (Ours)}& \checkmark&0\%&\textbf{60.01} &\textbf{44.81} &\textbf{10.39} &\textbf{8.98} &\textbf{22.24} &\textbf{32.60}&\textbf{23.86}\\
\bottomrule
\specialrule{1pt}{1pt}{0pt}
\end{tabular}
}
\caption{\textbf{The main results on WOD \emph{val} set.} `3D Sup.' means the ratio of video sequences with 3D labels. `Tem.' means using a temporal object detector. $\dag$: trained with 1/3 frames. }
\vspace{-5pt}
\label{tab:main}
\end{table*}
\begin{table}[]
\centering
\resizebox{\linewidth}{!}{
\begin{tabular}{l|c|ccc}
\specialrule{1pt}{0pt}{1pt}
\toprule
        Method & Extra Data & Easy & Mod. & Hard \\ 
        \midrule
        PatchNet~\citep{ma2020rethinking} & Raw+depth & 15.68 & 11.12 & 10.17 \\ 
        PCT~\citep{pct} & - & 21.00 & 13.37 & 11.31 \\
        GUPNet~\citep{lu2021gup} & - & 22.26 & 15.02 & 13.12 \\ 
        MonoDTR~\citep{huang2022monodtr} & - & 21.99 & 15.39 & 12.73 \\ 
        DCD~\citep{li2022densely}  & CAD models & 23.81 & 15.90 & 13.21 \\
        MonoJSG~\citep{lian2022monojsg}  & - & 24.69 & 16.14 & 13.64 \\ 
        DID-M3D~\citep{peng2022did} & - & \textbf{24.40} & 16.29 & \textbf{13.75} \\ 
        \textcolor{gray}{LPCG~\citep{peng2022lidar}} & \textcolor{gray}{Raw+LiDAR} & \textcolor{gray}{25.56} & \textcolor{gray}{17.80} & \textcolor{gray}{15.38} \\
        \textcolor{gray}{CMKD~\citep{hong2022cross}}  & \textcolor{gray}{Raw+LiDAR} & \textcolor{gray}{28.55} & \textcolor{gray}{18.69} & \textcolor{gray}{16.77} \\ 
        \midrule
        MonoFlex~\citep{zhang2021objects}  & - & 19.94 & 13.89 & 12.07 \\ 
        \textbf{BA$^2$-Det+MonoFlex (Ours)} & Raw & 23.45 & \textbf{16.30} & 13.50 \\ 
        \textcolor{dgreen}{\textit{Improvement}}& &\textcolor{dgreen}{\textit{+3.51}}&\textcolor{dgreen}{\textit{+2.41}} &\textcolor{dgreen}{\textit{+1.43}}\\
        \bottomrule
        \specialrule{1pt}{1pt}{0pt}
\end{tabular}}
\caption{\textbf{The results on \emph{test} set of KITTI detection benchmark.} `Raw' means using images in KITTI Raw set. `Mod.' means the moderate level of difficulty.}
\vspace{-10pt}
\label{tab:kitti}
\end{table}
To further refine the generalized object boxes, we iteratively retrain the detector with predictions as the updated pseudo labels. We adopt a retraining strategy of using 3D pseudo labels for initial training and updating labels with predictions from the last iteration 
\begin{equation}
\left\{\begin{aligned}
\mathcal{D}^{(0)}(X, Y) &= N_\theta^{(0)}(\widetilde{\mathcal{D}}(X, Y)),\\
\mathcal{D}^{(l)}(X, Y) &= N_\theta^{(l)}(\mathcal{D}^{(l-1)}(X, Y)),
\end{aligned}\right.
\label{eq:strategy2}
\end{equation}
where $\widetilde{\mathcal{D}}(X, Y)$ is the dataset with generated pseudo labels, $\mathcal{D}^{(l)}(X, Y)$ is the dataset with predicted labels, $(l)$ means the $l$-th self-training iteration. Note that we do not keep the last network parameters for each self-retraining iteration and train the network $N_\theta$ for the same $\kappa$ epochs.\par

\paragraph{Generalization from close to distant objects.}
A natural challenge for the monocular 3D object detector is that distant objects have fewer pixels, making it difficult to estimate their 3D position, especially depth estimation.
Inspired by BA-Det~\citep{he2022badet}, we utilize geometric temporal aggregation to generalize the close objects to distant objects. 
Specifically, we learn the object-centric feature correspondence for an object with the object-centric temporal correspondence learning (OTCL) module and solve object-centric bundle adjustment (OBA) between the tracked object prediction $\{\widetilde{\mathbf{B}}_j^1,\widetilde{\mathbf{B}}_j^2,\cdots,\widetilde{\mathbf{B}}_j^T\}$ from $N_\theta$ during inference.\par
\section{Experiments}
\begin{table*}[]
\centering
\resizebox{0.8\linewidth}{!}{
\begin{tabular}{c|cccccc|cccc}
\specialrule{1pt}{0pt}{1pt}
\toprule
&3D Ass. &2D Ass. &$G_\theta$ &Ori. & Iter. &OBA &3D AP$_\text{5}$ & 3D APH$_\text{5}$ &LET APL$_\text{50}$ & LET AP$_\text{50}$ \\
\midrule
(a)&\checkmark & & & & &&20.97 &6.70 &4.27 &7.28\\
(b)&&\checkmark &&&&&28.40 &11.34&5.02&8.62\\
\midrule
(c)&\cellcolor{gray!15}&\cellcolor{gray!15}\checkmark &\cellcolor{gray!15}\checkmark&\cellcolor{gray!15}&\cellcolor{gray!15}&\cellcolor{gray!15}&\cellcolor{gray!15}33.75 &\cellcolor{gray!15}11.94 &\cellcolor{gray!15}9.63 &\cellcolor{gray!15}16.80\\
(d)&\cellcolor{gray!30}&\cellcolor{gray!30}\checkmark &\cellcolor{gray!30}\checkmark &\cellcolor{gray!30}\checkmark&\cellcolor{gray!30}&\cellcolor{gray!30}&\cellcolor{gray!30}41.17 &\cellcolor{gray!30}28.73&\cellcolor{gray!30}12.23 &\cellcolor{gray!30}21.41\\
(e)&\cellcolor{gray!30}&\cellcolor{gray!30}\checkmark &\cellcolor{gray!30}\checkmark &\cellcolor{gray!30}\checkmark &\cellcolor{gray!30}\checkmark &\cellcolor{gray!30}&\cellcolor{gray!30}56.33 &\cellcolor{gray!30}42.05&\cellcolor{gray!30}17.87 &\cellcolor{gray!30}29.62\\
(f)&\cellcolor{gray!45}&\cellcolor{gray!45}\checkmark &\cellcolor{gray!45}\checkmark &\cellcolor{gray!45}\checkmark &\cellcolor{gray!45}\checkmark&\cellcolor{gray!45}\checkmark &\cellcolor{gray!45}\textbf{60.01} &\cellcolor{gray!45}\textbf{44.81} &\cellcolor{gray!45}\textbf{22.24} &\cellcolor{gray!45}\textbf{32.60}\\
\bottomrule
\specialrule{1pt}{1pt}{0pt}
\end{tabular}
}
\caption{\textbf{The ablation study on \methodname{}.}The gray cells mean \colorbox{gray!15}{complete to partial}, \colorbox{gray!30}{static to dynamic}, \colorbox{gray!45}{close to distant} generalization stages respectively.}
\vspace{-5pt}
\label{tab:ablate}
\end{table*}
\subsection{Datasets and metrics}
\paragraph{Waymo open dataset (WOD).} To verify our proposed \methodname{}, we conduct our ablation studies and comparison experiments with other methods on the large-scale autonomous driving dataset, Waymo Open Dataset (WOD)~\citep{sun2020scalability}. WOD is the mainstream 3D object detection benchmark, containing 1150 video sequences, 798 for training, 202 for validation, and 150 for testing. Only objects within 75m that can be scanned by LiDAR have 3D labels.
To keep the same experiment settings as other methods, we mainly report the results on WOD v1.2. The evaluation metrics for camera-based 3D object detection are 3D AP and LET-3D AP~\citep{hung2022let}. 3D AP is a common metric for both camera and LiDAR-based 3D object detection. LET-3D AP is specifically designed for camera-only 3D object detection. Because the camera-based 3D object detector has a natural weakness in depth estimation, LET-3D AP is much looser for longitudinal localization and uses Longitudinal Error Tolerant IoU (LET-IoU) instead of the original IoU as the criterion. Following the existing camera-based 3D object detection methods, we mainly report the results of the VEHICLE class on the FRONT camera. For 3D AP and 3D APH, we choose a loose IoU threshold of 0.05 and a common one of 0.5, called AP$_\text{5}$ and AP$_\text{50}$. For LET-3D metrics, we report the results under the official IoU threshold of 0.5.\par
\paragraph{KITTI dataset.} 
 KITTI object detection benchmark consists of 7481 images for training and 7518 images for testing. Unlike WOD, it is not organized as long video sequences. The main evaluation metric is 3D AP on three difficulty levels, easy, moderate, and hard. Besides the object detection benchmark, KITTI also provides the raw dataset without 3D object labels. We train \methodname{} on KITTI raw dataset.
\subsection{Implementation Details}
\label{sec:impl}
For 3D object detector, we use a DLA-34~\citep{yu2018deep} as the backbone without an FPN neck, and the head is with 2 layers of 3$\times$3 convolutions and MLP. The resolution of the input images is 1920$\times$1280. If the input size is smaller than it, we will use zero padding to complete the image.
The scene reconstruction is based on hloc~\citep{sarlin2019coarse} framework. The speed filtering threshold $\omega$ is 1m/s. The distance threshold $\delta_1$ and $\delta_2$ in DoubleClustering are 0.5 and 0.7. We keep the object cluster for more than $\theta=100$ points. The size threshold $\sigma_0=3m$ and $\sigma_1=10m$. Our implementation is based on the PyTorch~\citep{paszke2019pytorch} framework. We train our model on 8 NVIDIA RTX 4090 GPUs. Adam~\citep{kingma2014adam} optimizer is applied with $\beta_1=0.9$ and $\beta_2=0.999$. The learning rate is 8$\times$10$^{-5}$ and weight decay is 10$^{-5}$. We train 1 epoch for network $G_\theta$ and $\kappa=12$ epochs for the 3D object detector $N_\theta$. The loss weights are the same as BA-Det. The self-retrain iteration number for $N_\theta$ is 2. As shown in Table~\ref{tab:depthablate}, we choose labels in the wider depth range $[0.5\text{m},200\text{m}]$ for initial training and regular range $[0.5\text{m},75\text{m}]$ for iterative self-retraining.\par
\subsection{Main Results}
As shown in Table~\ref{tab:main}, we show the main results compared with fully-supervised BA-Det, a simple solution combining SfM and BA-Det, and some other fully supervised methods. Especially, to make a clear understanding of our results, we also compare the results of BA-Det trained with fewer data. We find that with a loose IoU threshold (0.05), our \methodname{} can outperform fully-supervised BA-Det with 10\% data by 2.7 AP and is close to the 100\% data (with a $\sim$12 AP gap). As for the 0.5 IoU threshold, we can beat some other fully-supervised methods, such as PCT and MonoJSG. Compared with our baseline method, using SfM~\citep{schonberger2016structure} and clustering to fit 3D labels and learning a BA-Det, our \methodname{} have huge gains on all metrics. Besides, we also conduct experiments on the semi-supervised setting with the same 10\% GT as BA-Det and MonoFlex and show better performance.\par
We also conduct experiments on KITTI~\cite{geiger2012we} to compare with other SOTA methods. We report 3D object detection results on test set. Note that
(1) there is no long video for 3D reconstruction on KITTI object detection dataset, and we have to generate 3D pseudo boxes on KITTI raw dataset;
(2) there are hardly any comparable methods available for 2D supervised 3D detection.
So, we report the results taking \methodname{} as the pretraining approach. We pretrain \methodname{} on KITTI raw dataset without any labels (We use the 2D object detector Mask R-CNN trained on COCO) and finetune the monocular 3D object detector MonoFlex with 3D ground truth on KITTI detection training set. We only use a single frame during inference for a fair comparison. The results have been shown in Table~\ref{tab:kitti}. \methodname{} can have 2.4 AP gain (about 20\% relative improvement) on the moderate level.\par
\begin{table}[]
\centering
\resizebox{\columnwidth}{!}{
\begin{tabular}{c|cccc}
\toprule
Iter. & 3D AP$_\text{5}$ & 3D APH$_\text{5}$ & LET APL$_\text{50}$ & LET AP$_\text{50}$ \\
\midrule
0&41.17 &28.73&12.23 &21.41\\
1&51.81 &38.66&17.42 &28.57\\
2&56.33 &42.05&\textbf{17.87} &\textbf{29.62}\\
3&56.70 &42.27 &16.87 &28.06\\
4 &\textbf{57.12} &\textbf{42.42} &16.64 &27.73\\
\bottomrule
\end{tabular}
}
\caption{\textbf{Discussion about self-retraining iterations.}}
\label{tab:retrainiter}
\vspace{-15pt}
\end{table}
\subsection{Ablation Study}
\label{sec:ablate}
We conduct the ablation study on WOD val set. Results are shown in Table~\ref{tab:ablate}. 
(a) `3D Ass.' and (b) `2D Ass.' mean we assign labels based on 3D or 2D (pseudo) labels (Fig.~\ref{fig:label_assign}). 3D assignment will keep the recall and enhance the refinement of the 3D position during the iterative self-retraining stage. (c) $G_\theta$ learns a complete 3D box from the partially reconstructed object, that can obtain a 5.4 AP gain. (d) `Ori.' is the orientation optimization, improving 16.8 APH by refining the orientation and box size. (e) Iterative self-retraining (Iter.) has the greatest gain of 15.2 AP. Comparing (e) and (f), with temporal OBA, 3D bounding boxes can generalize from near to far, which improves 3.7 AP.  For more ablation studies, please refer to the appendix.

\begin{table*}[t]

   \centering
    \resizebox{0.85\linewidth}{!}{
        \begin{tabular}{c|c|ccc|cccc}
            \toprule
                         &Ratio    &$\delta< 1.25\uparrow$  &$\delta< 1.25^2\uparrow$ 
&$\delta < 1.25^3\uparrow$  &Abs Rel$\downarrow$ &Sq Rel$\downarrow$ &RMSE$\downarrow$ &RMSE log$\downarrow$      \\
            \midrule
            w/o p.l.&52.2\% &0.991&0.994&0.995&0.055&0.324&2.773&0.100\\
            w/ p.l.&47.8\% &0.995&0.996&0.997&0.049&0.117&1.726&0.077\\
            \midrule
            All&100\% &0.992&0.995&0.996&0.053&0.266&2.524&0.094\\
            \bottomrule
        \end{tabular}
    }
        \caption{\textbf{Satic-to-dynamic generalization ability of 3D localization on WOD training set.} `w/o p.l.': objects without pseudo label, `w/ p.l.': objects with pseudo label.}
             \vspace{-5pt}
    \label{tab:dynobj}
\end{table*}

\subsection{Discussions}
\label{sec:disc}
\paragraph{Self-retraining iteration.}
We discuss the number of iterations required for the self-retraining stage. In Table~\ref{tab:retrainiter}, we conduct the experiments of a maximum of 4 iterations. The first iteration can bring the greatest benefit of 10.7 AP and the benefits are progressively decreased. After the second iteration, the performance is near the highest performance, which shows fast convergence of the self-retraining stage. \par
\vspace{-5pt}
\paragraph{Retraining strategy.}
\begin{table}[]

\centering
\resizebox{\columnwidth}{!}{
\begin{tabular}{c|cccc}
\toprule
Strategy & 3D AP$_\text{5}$ & 3D APH$_\text{5}$ &LET APL$_\text{50}$  & LET AP$_\text{50}$ \\
\midrule
w/o iter. &41.17 &28.73&12.23 &21.41\\
\midrule
Eq.~\ref{eq:strategy1}&39.71\down{1.5}&29.79\up{1.0}&12.75\up{0.5}&21.08\down{0.3}\\
Eq.~\ref{eq:strategy2}&51.81\up{10.6} &38.66\up{9.9}&17.42\up{5.2} &28.57\up{7.2}\\
\bottomrule
\end{tabular}}
\caption{\textbf{Ablation study on self-retraining strategy.}}
\label{tab:retrainabl}
\end{table}
In the static to dynamic generalization stage, we propose the iterative retraining strategy (Eq.~\ref{eq:strategy2}).
Another feasible retraining strategy can be keeping the initial 3D pseudo labels and supplementing the high score predictions iteratively
\begin{equation}
    \mathcal{D}^{(l)}(X, Y) = N_\theta^{(l)}(\widetilde{\mathcal{D}}(X, Y)\cup \mathcal{D}^{(l-1)}(X, Y)),
\label{eq:strategy1}
\end{equation}
 In Table~\ref{tab:retrainabl}, we show the experiment results of retaining 1 iteration. The strategy in Eq.~\ref{eq:strategy2} is much better. The combination of pseudo-labels and predictions may introduce two inconsistent data distributions, causing poor results. The results in Table~\ref{tab:ablate} and Table~\ref{tab:retrainiter} also indicate the importance of the chosen self-retraining strategy.\par
\begin{figure}[h]
	\centering
        \includegraphics[width =0.8\linewidth]{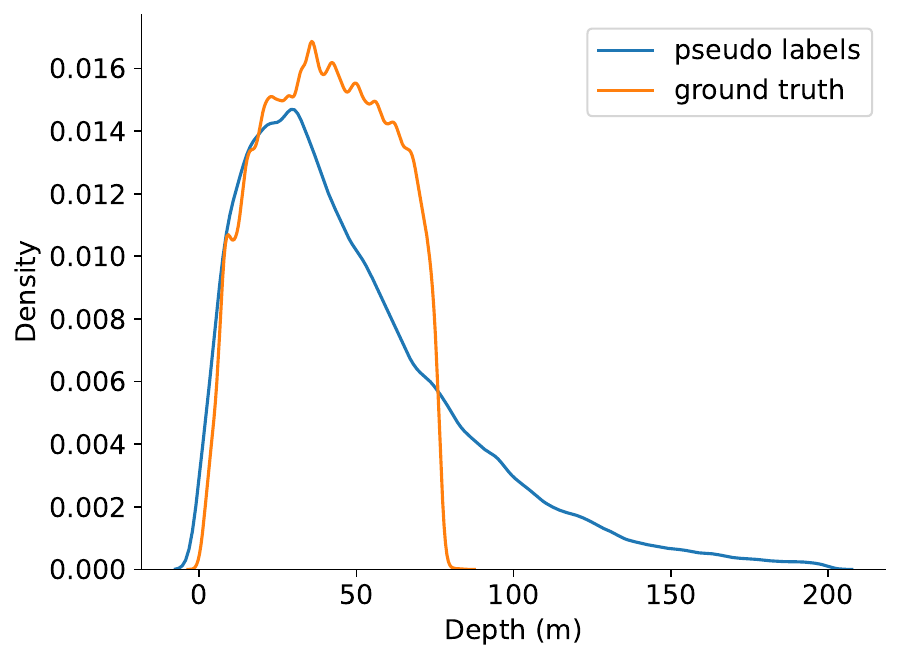}
        \vspace{-5pt}
	\caption{\textbf{Depth distributions of ground truth and pseudo labels.} \methodname{} can generate pseudo labels up to a maximum of 200m, whereas the ground truth labels range from 0m to 75m.}
        \vspace{-10pt}
	\label{fig:depth}
\end{figure}

\vspace{-5pt}
\paragraph{Depth threshold settings.}
\begin{table}[]
\centering
\resizebox{\columnwidth}{!}{
\begin{tabular}{c|c|cccc}
\toprule
Iter. &Depth (m) & 3D AP$_\text{5}$ & 3D APH$_\text{5}$ & LET APL$_\text{50}$ & LET AP$_\text{50}$ \\
\midrule
0&$[0.5,75]$ &22.24 &7.66 &4.83 &8.56\\
0&all, $[0.5,200]$&\textbf{33.75}&\textbf{11.94}&\textbf{9.63}&\textbf{16.80}\\
\midrule
1&$[0.5,75]$ &\textbf{42.07}&\textbf{15.00}&\textbf{12.32}&\textbf{20.48}\\
1&all, $[0.5,200]$&40.60&14.49&12.18&20.05\\
\bottomrule
\end{tabular}}
\caption{\textbf{Experiments for different depth thresholds.}}
\vspace{-10pt}
\label{tab:depthablate}
\end{table}
By utilizing reconstruction, we can generate labels for objects that are farther away, and thus the depth distribution is different from ground truth, shown in Fig.~\ref{fig:depth}. According to experiments (Table~\ref{tab:depthablate}), we find that when we initially train the object detector, we need these farther objects, and for iteratively self-retraining, we can only train with the same depth ranges as the ground truth (0-75m).\par
\vspace{-5pt}
\paragraph{3D location results for static to dynamic generalization.}
In Table~\ref{tab:dynobj}, we validate the performance of generalizing 3D location for unlabeled objects. The metrics follow depth estimation but on the object level. We find that only 47.8\% objects can be generated 3D pseudo labels directly. The others are mostly moving objects. The small gap between labeled and unlabeled objects in performance shows the static-to-dynamic generalization of the network $N_\theta$.
\vspace{-10pt}
\paragraph{Open-set 3D object detection, 3D tracking results, more ablation studies, discussions, qualitative results, limitations, and broader impacts.} Please refer to the appendix.  For video demos, please refer to \url{https://ba2det.site}.

\vspace{-5pt}
\section{Conclusion}
In this paper, we propose \methodname{}, a novel paradigm for 2D supervised monocular 3D object detection. The key idea of \methodname{} is to generate 3D pseudo labels from scene-level global reconstruction and learn neural networks to generalize pseudo labels in multi-stage. Specifically, in pseudo label generation phase, objects are clustered by the proposed DoubleClustering algorithm and fitted into bounding boxes. In multi-stage generalization phase, we design three stages of generalization: progressing from complete to partial, static to dynamic, and near to far.
 Experiments on the large-scale Waymo Open Dataset show that the performance of \methodname{} is on par with the fully-supervised BA-Det trained with 10\% videos and even outperforms some pioneering fully-supervised methods. As a pretraining method, \methodname{} can bring 20\% relative improvement on KITTI dataset.
We also show the potential of \methodname{} for detecting open-set 3D objects in complex scenes.\par

\bibliography{imcl2024}

\begin{thebibliography}{54}
\providecommand{\natexlab}[1]{#1}
\providecommand{\url}[1]{\texttt{#1}}
\expandafter\ifx\csname urlstyle\endcsname\relax
  \providecommand{\doi}[1]{doi: #1}\else
  \providecommand{\doi}{doi: \begingroup \urlstyle{rm}\Url}\fi

\bibitem[Brazil \& Liu(2019)Brazil and Liu]{brazil2019m3d}
Brazil, G. and Liu, X.
\newblock M3d-rpn: Monocular 3d region proposal network for object detection.
\newblock In \emph{ICCV}, 2019.

\bibitem[Campello et~al.(2013)Campello, Moulavi, and Sander]{campello2013density}
Campello, R.~J., Moulavi, D., and Sander, J.
\newblock Density-based clustering based on hierarchical density estimates.
\newblock In \emph{KDD}, 2013.

\bibitem[Chen et~al.(2016)Chen, Kundu, Zhang, Ma, Fidler, and Urtasun]{chen2016monocular}
Chen, X., Kundu, K., Zhang, Z., Ma, H., Fidler, S., and Urtasun, R.
\newblock Monocular 3d object detection for autonomous driving.
\newblock In \emph{CVPR}, 2016.

\bibitem[DeTone et~al.(2018)DeTone, Malisiewicz, and Rabinovich]{detone2018superpoint}
DeTone, D., Malisiewicz, T., and Rabinovich, A.
\newblock Superpoint: Self-supervised interest point detection and description.
\newblock In \emph{CVPR Workshops}, 2018.

\bibitem[Ding et~al.(2020)Ding, Huo, Yi, Wang, Shi, Lu, and Luo]{ding2020learning}
Ding, M., Huo, Y., Yi, H., Wang, Z., Shi, J., Lu, Z., and Luo, P.
\newblock Learning depth-guided convolutions for monocular 3d object detection.
\newblock In \emph{CVPR}, 2020.

\bibitem[Fan et~al.(2023)Fan, Yang, Wang, Wang, and Zhang]{fan2023super}
Fan, L., Yang, Y., Wang, F., Wang, N., and Zhang, Z.
\newblock Super sparse 3d object detection.
\newblock \emph{arXiv preprint arXiv:2301.02562}, 2023.

\bibitem[Fischer et~al.(2022)Fischer, Yang, Kumar, Sun, and Yu]{fischer2022cc}
Fischer, T., Yang, Y.-H., Kumar, S., Sun, M., and Yu, F.
\newblock Cc-3dt: Panoramic 3d object tracking via cross-camera fusion.
\newblock \emph{arXiv preprint arXiv:2212.01247}, 2022.

\bibitem[Geiger et~al.(2012)Geiger, Lenz, and Urtasun]{geiger2012we}
Geiger, A., Lenz, P., and Urtasun, R.
\newblock Are we ready for autonomous driving? the kitti vision benchmark suite.
\newblock In \emph{CVPR}, 2012.

\bibitem[He et~al.(2023)He, Chen, Wang, and Zhang]{he2022badet}
He, J., Chen, Y., Wang, N., and Zhang, Z.
\newblock 3d video object detection with learnable object-centric global optimization.
\newblock In \emph{CVPR}, 2023.

\bibitem[Hong et~al.(2022)Hong, Dai, and Ding]{hong2022cross}
Hong, Y., Dai, H., and Ding, Y.
\newblock Cross-modality knowledge distillation network for monocular 3d object detection.
\newblock In \emph{ECCV}, 2022.

\bibitem[Hu et~al.(2022)Hu, Yang, Fischer, Darrell, Yu, and Sun]{hu2022monocular}
Hu, H.-N., Yang, Y.-H., Fischer, T., Darrell, T., Yu, F., and Sun, M.
\newblock Monocular quasi-dense 3d object tracking.
\newblock \emph{IEEE Transactions on Pattern Analysis and Machine Intelligence}, 45\penalty0 (2):\penalty0 1992--2008, 2022.

\bibitem[Huang et~al.(2022)Huang, Wu, Su, and Hsu]{huang2022monodtr}
Huang, K.-C., Wu, T.-H., Su, H.-T., and Hsu, W.~H.
\newblock Monodtr: Monocular 3d object detection with depth-aware transformer.
\newblock In \emph{CVPR}, 2022.

\bibitem[Hung et~al.(2022)Hung, Kretzschmar, Casser, Hwang, and Anguelov]{hung2022let}
Hung, W.-C., Kretzschmar, H., Casser, V., Hwang, J.-J., and Anguelov, D.
\newblock Let-3d-ap: Longitudinal error tolerant 3d average precision for camera-only 3d detection.
\newblock \emph{arXiv preprint arXiv:2206.07705}, 2022.

\bibitem[Kingma \& Ba(2014)Kingma and Ba]{kingma2014adam}
Kingma, D.~P. and Ba, J.
\newblock Adam: A method for stochastic optimization.
\newblock \emph{arXiv preprint arXiv:1412.6980}, 2014.

\bibitem[Kirillov et~al.(2023)Kirillov, Mintun, Ravi, Mao, Rolland, Gustafson, Xiao, Whitehead, Berg, Lo, et~al.]{kirillov2023segment}
Kirillov, A., Mintun, E., Ravi, N., Mao, H., Rolland, C., Gustafson, L., Xiao, T., Whitehead, S., Berg, A.~C., Lo, W.-Y., et~al.
\newblock Segment anything.
\newblock \emph{arXiv preprint arXiv:2304.02643}, 2023.

\bibitem[Li et~al.(2022{\natexlab{a}})Li, Shugurov, Busam, Yang, and Ilic]{li2022ws}
Li, F., Shugurov, I., Busam, B., Yang, S., and Ilic, S.
\newblock Ws-ope: Weakly supervised 6-d object pose regression using relative multi-camera pose constraints.
\newblock \emph{IEEE Robotics and Automation Letters}, 7\penalty0 (2):\penalty0 3703--3710, 2022{\natexlab{a}}.

\bibitem[Li et~al.(2022{\natexlab{b}})Li, Chen, He, and Zhang]{li2022densely}
Li, Y., Chen, Y., He, J., and Zhang, Z.
\newblock Densely constrained depth estimator for monocular 3d object detection.
\newblock In \emph{ECCV}, 2022{\natexlab{b}}.

\bibitem[Lian et~al.(2022)Lian, Li, and Chen]{lian2022monojsg}
Lian, Q., Li, P., and Chen, X.
\newblock Monojsg: Joint semantic and geometric cost volume for monocular 3d object detection.
\newblock In \emph{CVPR}, 2022.

\bibitem[Liu et~al.(2021)Liu, Zhang, Cao, Hu, and Tong]{liu2021group}
Liu, Z., Zhang, Z., Cao, Y., Hu, H., and Tong, X.
\newblock Group-free 3d object detection via transformers.
\newblock In \emph{ICCV}, 2021.

\bibitem[Lu et~al.(2021)Lu, Ma, Yang, Zhang, Liu, Chu, Yan, and Ouyang]{lu2021gup}
Lu, Y., Ma, X., Yang, L., Zhang, T., Liu, Y., Chu, Q., Yan, J., and Ouyang, W.
\newblock Geometry uncertainty projection network for monocular 3d object detection.
\newblock In \emph{ICCV}, 2021.

\bibitem[Ma et~al.(2020)Ma, Liu, Xia, Zhang, Zeng, and Ouyang]{ma2020rethinking}
Ma, X., Liu, S., Xia, Z., Zhang, H., Zeng, X., and Ouyang, W.
\newblock Rethinking pseudo-lidar representation.
\newblock In \emph{ECCV}, 2020.

\bibitem[Mousavian et~al.(2017)Mousavian, Anguelov, Flynn, and Kosecka]{mousavian20173d}
Mousavian, A., Anguelov, D., Flynn, J., and Kosecka, J.
\newblock 3d bounding box estimation using deep learning and geometry.
\newblock In \emph{CVPR}, 2017.

\bibitem[Najibi et~al.(2022)Najibi, Ji, Zhou, Qi, Yan, Ettinger, and Anguelov]{najibi2022motion}
Najibi, M., Ji, J., Zhou, Y., Qi, C.~R., Yan, X., Ettinger, S., and Anguelov, D.
\newblock Motion inspired unsupervised perception and prediction in autonomous driving.
\newblock In \emph{ECCV}, 2022.

\bibitem[Nunes et~al.(2022)Nunes, Chen, Marcuzzi, Osep, Leal-Taix{\'e}, Stachniss, and Behley]{nunes2022unsupervised}
Nunes, L., Chen, X., Marcuzzi, R., Osep, A., Leal-Taix{\'e}, L., Stachniss, C., and Behley, J.
\newblock Unsupervised class-agnostic instance segmentation of 3d lidar data for autonomous vehicles.
\newblock \emph{IEEE Robotics and Automation Letters}, 7\penalty0 (4):\penalty0 8713--8720, 2022.

\bibitem[Paszke et~al.(2019)Paszke, Gross, Massa, Lerer, Bradbury, Chanan, Killeen, Lin, Gimelshein, Antiga, et~al.]{paszke2019pytorch}
Paszke, A., Gross, S., Massa, F., Lerer, A., Bradbury, J., Chanan, G., Killeen, T., Lin, Z., Gimelshein, N., Antiga, L., et~al.
\newblock Pytorch: An imperative style, high-performance deep learning library.
\newblock \emph{NeurIPS}, 2019.

\bibitem[Peng et~al.(2022{\natexlab{a}})Peng, Liu, Yu, Yan, Deng, Yang, Liu, and Cai]{peng2022lidar}
Peng, L., Liu, F., Yu, Z., Yan, S., Deng, D., Yang, Z., Liu, H., and Cai, D.
\newblock Lidar point cloud guided monocular 3d object detection.
\newblock In \emph{ECCV}, 2022{\natexlab{a}}.

\bibitem[Peng et~al.(2022{\natexlab{b}})Peng, Wu, Yang, Liu, and Cai]{peng2022did}
Peng, L., Wu, X., Yang, Z., Liu, H., and Cai, D.
\newblock Did-m3d: Decoupling instance depth for monocular 3d object detection.
\newblock In \emph{ECCV}, 2022{\natexlab{b}}.

\bibitem[Peng et~al.(2022{\natexlab{c}})Peng, Yan, Wu, Yang, He, and Cai]{peng2022weakm3d}
Peng, L., Yan, S., Wu, B., Yang, Z., He, X., and Cai, D.
\newblock Weakm3d: Towards weakly supervised monocular 3d object detection.
\newblock \emph{arXiv preprint arXiv:2203.08332}, 2022{\natexlab{c}}.

\bibitem[Qi et~al.(2019)Qi, Litany, He, and Guibas]{qi2019deep}
Qi, C.~R., Litany, O., He, K., and Guibas, L.~J.
\newblock Deep hough voting for 3d object detection in point clouds.
\newblock In \emph{ICCV}, 2019.

\bibitem[Qian et~al.(2020)Qian, Garg, Wang, You, Belongie, Hariharan, Campbell, Weinberger, and Chao]{qian2020end}
Qian, R., Garg, D., Wang, Y., You, Y., Belongie, S., Hariharan, B., Campbell, M., Weinberger, K.~Q., and Chao, W.-L.
\newblock End-to-end pseudo-lidar for image-based 3d object detection.
\newblock In \emph{CVPR}, 2020.

\bibitem[Reading et~al.(2021)Reading, Harakeh, Chae, and Waslander]{reading2021categorical}
Reading, C., Harakeh, A., Chae, J., and Waslander, S.~L.
\newblock Categorical depth distribution network for monocular 3d object detection.
\newblock In \emph{CVPR}, 2021.

\bibitem[Sarlin et~al.(2019)Sarlin, Cadena, Siegwart, and Dymczyk]{sarlin2019coarse}
Sarlin, P.-E., Cadena, C., Siegwart, R., and Dymczyk, M.
\newblock From coarse to fine: Robust hierarchical localization at large scale.
\newblock In \emph{CVPR}, 2019.

\bibitem[Sarlin et~al.(2020)Sarlin, DeTone, Malisiewicz, and Rabinovich]{sarlin2020superglue}
Sarlin, P.-E., DeTone, D., Malisiewicz, T., and Rabinovich, A.
\newblock Superglue: Learning feature matching with graph neural networks.
\newblock In \emph{CVPR}, 2020.

\bibitem[Schonberger \& Frahm(2016)Schonberger and Frahm]{schonberger2016structure}
Schonberger, J.~L. and Frahm, J.-M.
\newblock Structure-from-motion revisited.
\newblock In \emph{CVPR}, 2016.

\bibitem[Shi et~al.(2020)Shi, Guo, Jiang, Wang, Shi, Wang, and Li]{shi2020pv}
Shi, S., Guo, C., Jiang, L., Wang, Z., Shi, J., Wang, X., and Li, H.
\newblock Pv-rcnn: Point-voxel feature set abstraction for 3d object detection.
\newblock In \emph{CVPR}, 2020.

\bibitem[Sun et~al.(2020)Sun, Kretzschmar, Dotiwalla, Chouard, Patnaik, Tsui, Guo, Zhou, Chai, Caine, et~al.]{sun2020scalability}
Sun, P., Kretzschmar, H., Dotiwalla, X., Chouard, A., Patnaik, V., Tsui, P., Guo, J., Zhou, Y., Chai, Y., Caine, B., et~al.
\newblock Scalability in perception for autonomous driving: Waymo open dataset.
\newblock In \emph{CVPR}, 2020.

\bibitem[Triebel et~al.(2010)Triebel, Shin, and Siegwart]{triebel2010segmentation}
Triebel, R., Shin, J., and Siegwart, R.
\newblock Segmentation and unsupervised part-based discovery of repetitive objects.
\newblock In \emph{RSS}, 2010.

\bibitem[Wang et~al.(2021{\natexlab{a}})Wang, Zhang, Zhu, Zhang, He, Li, and Xue]{pct}
Wang, L., Zhang, L., Zhu, Y., Zhang, Z., He, T., Li, M., and Xue, X.
\newblock Progressive coordinate transforms for monocular 3d object detection.
\newblock \emph{NeurIPS}, 2021{\natexlab{a}}.

\bibitem[Wang et~al.(2021{\natexlab{b}})Wang, Chen, Pang, Wang, and Zhang]{wang2021immortal}
Wang, Q., Chen, Y., Pang, Z., Wang, N., and Zhang, Z.
\newblock Immortal tracker: Tracklet never dies.
\newblock \emph{arXiv preprint arXiv:2111.13672}, 2021{\natexlab{b}}.

\bibitem[Wang et~al.(2023)Wang, Liu, Wang, Li, and Zhang]{Wang_2023_ICCV}
Wang, S., Liu, Y., Wang, T., Li, Y., and Zhang, X.
\newblock Exploring object-centric temporal modeling for efficient multi-view 3d object detection.
\newblock In \emph{ICCV}, 2023.

\bibitem[Wang et~al.(2021{\natexlab{c}})Wang, Zhu, Pang, and Lin]{wang2021fcos3d}
Wang, T., Zhu, X., Pang, J., and Lin, D.
\newblock Fcos3d: Fully convolutional one-stage monocular 3d object detection.
\newblock In \emph{ICCV}, 2021{\natexlab{c}}.

\bibitem[Wang et~al.(2022{\natexlab{a}})Wang, Pang, and Lin]{wang2022monocular}
Wang, T., Pang, J., and Lin, D.
\newblock Monocular 3d object detection with depth from motion.
\newblock In \emph{ECCV}, 2022{\natexlab{a}}.

\bibitem[Wang et~al.(2019)Wang, Chao, Garg, Hariharan, Campbell, and Weinberger]{wang2019pseudo}
Wang, Y., Chao, W.-L., Garg, D., Hariharan, B., Campbell, M., and Weinberger, K.~Q.
\newblock Pseudo-lidar from visual depth estimation: Bridging the gap in 3d object detection for autonomous driving.
\newblock In \emph{CVPR}, 2019.

\bibitem[Wang et~al.(2022{\natexlab{b}})Wang, Chen, and ZHANG]{wang20224d}
Wang, Y., Chen, Y., and ZHANG, Z.-X.
\newblock 4d unsupervised object discovery.
\newblock In \emph{NeurIPS}, 2022{\natexlab{b}}.

\bibitem[Yang et~al.(2022)Yang, Wang, Ge, Mao, Li, and Zhang]{yang2022towards}
Yang, J., Wang, T., Ge, Z., Mao, W., Li, X., and Zhang, X.
\newblock Towards 3d object detection with 2d supervision.
\newblock \emph{arXiv preprint arXiv:2211.08287}, 2022.

\bibitem[Yang et~al.(2021)Yang, Yu, and Yang]{yang2021dsc}
Yang, Z., Yu, X., and Yang, Y.
\newblock Dsc-posenet: Learning 6dof object pose estimation via dual-scale consistency.
\newblock In \emph{CVPR}, 2021.

\bibitem[Yin et~al.(2021)Yin, Zhou, and Krahenbuhl]{yin2021center}
Yin, T., Zhou, X., and Krahenbuhl, P.
\newblock Center-based 3d object detection and tracking.
\newblock In \emph{CVPR}, 2021.

\bibitem[You et~al.(2020)You, Wang, Chao, Garg, Pleiss, Hariharan, Campbell, and Weinberger]{You2020Pseudo-LiDAR++:}
You, Y., Wang, Y., Chao, W.-L., Garg, D., Pleiss, G., Hariharan, B., Campbell, M., and Weinberger, K.~Q.
\newblock Pseudo-lidar++: Accurate depth for 3d object detection in autonomous driving.
\newblock In \emph{ICLR}, 2020.

\bibitem[You et~al.(2022)You, Luo, Phoo, Chao, Sun, Hariharan, Campbell, and Weinberger]{you2022learning}
You, Y., Luo, K., Phoo, C.~P., Chao, W.-L., Sun, W., Hariharan, B., Campbell, M., and Weinberger, K.~Q.
\newblock Learning to detect mobile objects from lidar scans without labels.
\newblock In \emph{CVPR}, 2022.

\bibitem[Yu et~al.(2018)Yu, Wang, Shelhamer, and Darrell]{yu2018deep}
Yu, F., Wang, D., Shelhamer, E., and Darrell, T.
\newblock Deep layer aggregation.
\newblock In \emph{CVPR}, 2018.

\bibitem[Zakharov et~al.(2020)Zakharov, Kehl, Bhargava, and Gaidon]{zakharov2020autolabeling}
Zakharov, S., Kehl, W., Bhargava, A., and Gaidon, A.
\newblock Autolabeling 3d objects with differentiable rendering of sdf shape priors.
\newblock In \emph{CVPR}, 2020.

\bibitem[Zhang et~al.(2017)Zhang, Xu, Dong, and Dolan]{zhang2017efficient}
Zhang, X., Xu, W., Dong, C., and Dolan, J.~M.
\newblock Efficient l-shape fitting for vehicle detection using laser scanners.
\newblock In \emph{IV}, 2017.

\bibitem[Zhang et~al.(2021)Zhang, Lu, and Zhou]{zhang2021objects}
Zhang, Y., Lu, J., and Zhou, J.
\newblock Objects are different: Flexible monocular 3d object detection.
\newblock In \emph{CVPR}, 2021.

\bibitem[Zhou et~al.(2019)Zhou, Wang, and Kr{\"a}henb{\"u}hl]{zhou2019objects}
Zhou, X., Wang, D., and Kr{\"a}henb{\"u}hl, P.
\newblock Objects as points.
\newblock \emph{arXiv preprint arXiv:1904.07850}, 2019.

\end{thebibliography}
\bibliographystyle{icml2024}

\newpage
\appendix
\onecolumn

\appendix

\section{Additional Experiments}
\label{sec:sup-exp}

\subsection{3D Tracking Results}
\begin{table}[h]
   \begin{center}
    \resizebox{\linewidth}{!}{
        \begin{tabular}{l|c|cccc}
            \toprule
                                &Fully Sup.              & MOTA$_{50}$ $\uparrow$ & Mismatch$_{50}$ $\downarrow$  & MOTA$_{30}$ $\uparrow$ & Mismatch$_{30}$  $\downarrow$        \\
            \midrule
          QD-3DT~\citep{hu2022monocular} &\checkmark&0.0308 &0.00550 &0.1867 &0.01340  \\
          CC-3DT~\citep{fischer2022cc} &\checkmark&0.0480 &0.00180 &0.2032 &0.00690 \\
          \midrule
          SfM+BA-Det+Immortal~\citep{wang2021immortal} & &0.0011& $<$0.00001 &0.0652&0.00038\\
         \methodname{} (Ours) & &0.0352 &0.00002 &0.1522 &0.00008 \\
            \bottomrule
        \end{tabular}
    }
    \end{center}
    \caption{\textbf{Comparisons with SOTA methods on WOD for 3D MOT.}}
    \label{tab:mot3d}
\end{table}

In Table~\ref{tab:mot3d}, we show additional 3D MOT results on WOD. We compare the proposed \methodname{} with other SOTA monocular 3D MOT methods on WOD. All results are reported in Vehicle LEVEL 2 difficulty. 50 and 30 in the metrics are for the IoU threshold of 0.5 and 0.3. Note that the SOTA methods are both fully supervised, learning with 3D ground truth. Our \methodname{} is a 2D supervised 3D multiple object tracking method. Even though, \methodname{} is also comparable to these fully supervised ones. Especially for MOTA$_{50}$ metric, we outperform QD-3DT~\citep{hu2022monocular}. For Mismatch, we only have less than 1/100 identity switches compared with the SOTA methods. Compared with the baseline method, using SfM to generate pseudo labels, BA-Det as the 3D object detector, and ImmortalTracker to track 3D objects, our method \methodname{} improves the performance by a large margin. 
\subsection{Open-Set 3D Object Detection with SAM}
\label{sec:openset}
\begin{figure}[h]
          \begin{center}
          \resizebox{0.8\linewidth}{!}{
          	\subfloat[Input image examples.]{\includegraphics[width = 0.52\linewidth]{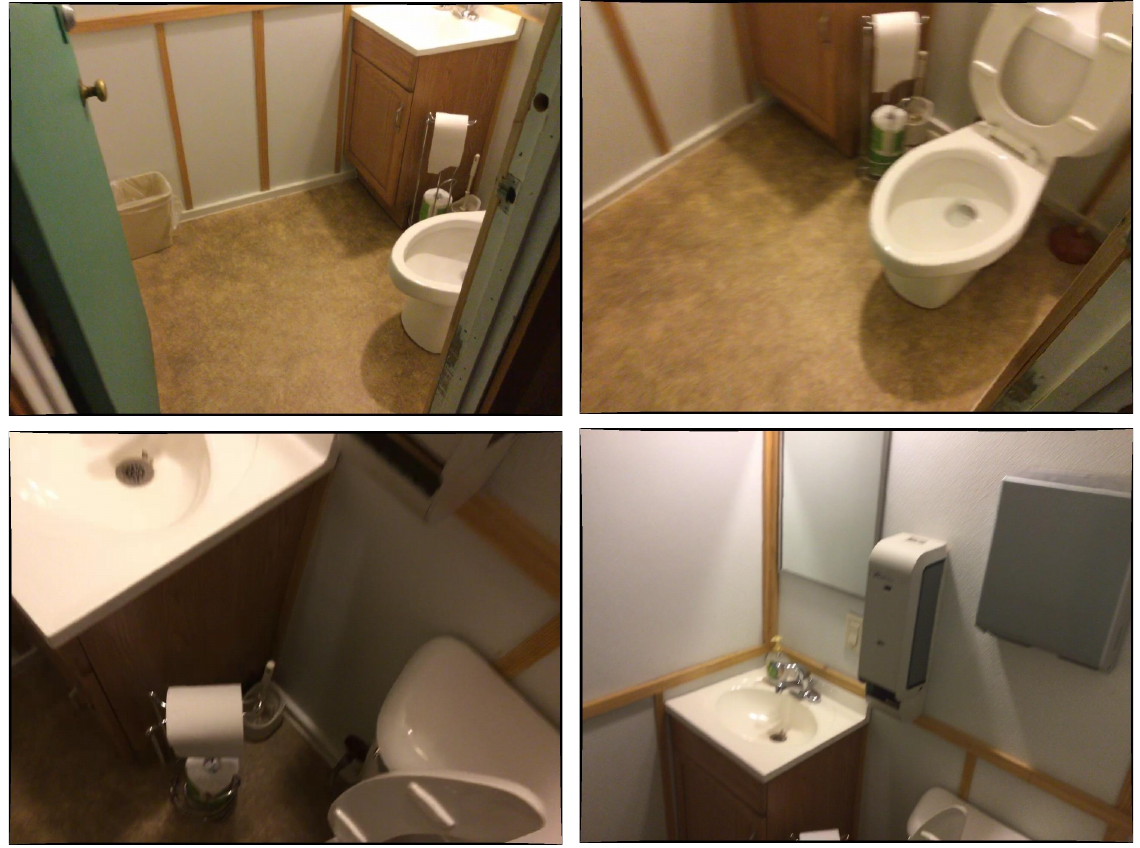}}
        	\hfill\ \ 
        	\subfloat[Detected 3D boxes.]{\includegraphics[width = 0.41\linewidth]{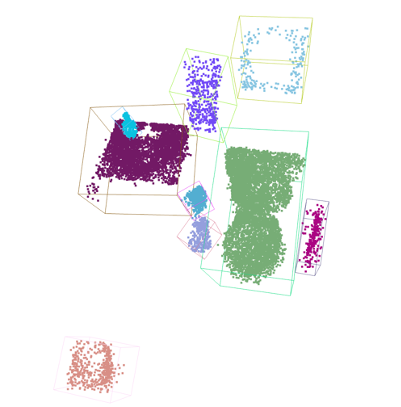}} 
              }
          \end{center}
          \vspace{-10pt}
             \caption{\textbf{Open-set 3D object detection from a video sequence.} For the video demos for 3D box generation, please refer to our project page.}
             \label{fig:openset}
\end{figure}

In Fig.~\ref{fig:openset}, we also show the ability to detect open-set 3D objects in complex scenes with SAM~\citep{kirillov2023segment} instead of 2D ground truth. We click the objects to generate the 2D masks in SAM. Please refer to the detailed video demos from the project page.
\subsection{Additional Ablation Studies and Discussions}
\subsubsection{Detailed results in different depth ranges.} The detailed results in different depth ranges are shown in Table~\ref{tab:range}. Compared with the baseline, we have a more significant gain for the objects far off.\par
\begin{table}[h]
\centering
\resizebox{\linewidth}{!}{
\begin{tabular}{l|c|ccc|ccc|ccc|ccc}
\specialrule{1pt}{0pt}{1pt}
\toprule
\multirow{2}{*}{Method}&\multirow{2}{*}{3D Sup.} &\multicolumn{3}{c|}{3D AP$_{5}$} &\multicolumn{3}{c|}{3D APH$_{5}$}&\multicolumn{3}{c|}{LET APL$_\text{50}$} &\multicolumn{3}{c}{LET AP$_\text{50}$} \\
  &&0-30 &30-50 &50-$\infty$ &0-30 &30-50 &50-$\infty$ &0-30 &30-50 &50-$\infty$ &0-30 &30-50 &50-$\infty$ \\ 
\midrule
BA-Det&100\%&87.80 &72.52 &48.45 &86.91 &71.52 &46.98 &66.15 &57.97 &36.44 &82.74 &69.58 &45.77 \\
 BA-Det&10\%&73.25 &54.00 &34.50 &71.38 &52.22 &32.53 &38.31&35.57&22.40&56.98&47.28&31.11\\
\midrule
 SfM+BA-Det&0\% &46.87 &25.88 &9.09 &14.26 &8.86 &2.84 &11.35 &7.74 &2.60 &17.59 &10.12 &3.48\\
 \methodname{} (Ours)&0\%&77.38 &54.95 &33.74 &64.54 &37.57 &21.64 &25.00 &23.97 &14.63 &39.24 &31.73 &20.30 \\
\bottomrule
\specialrule{1pt}{1pt}{0pt}
\end{tabular}}
\caption{\textbf{The detailed results in different depth ranges (meters) on WOD \emph{val} set.} }
\label{tab:range}
\end{table}

\subsubsection{Robustness across various levels of reconstruction quality.} Due to variations in reconstruction quality being primarily caused by differences in descriptors and matching algorithms, we simulate a scenario where there is a 25\% decrease in the number of matched points due to failed matching. The results are presented in Table~\ref{tab:rob_rec}. The results show that our method is robust to the worse reconstruction quality.\par
\begin{table}[h]
    \centering
    \begin{tabular}{l|cccc}
    \toprule
        &3D AP$_\text{5}$&3D APH$_\text{5}$&LET APL$_\text{50}$&LET AP$_\text{50}$ \\ 
        \midrule
        75\% points & 37.02 & 26.20 & 9.75 & 17.19 \\ 
        100\% points & 41.17 & 28.73 & 12.23 & 21.41 \\
        \bottomrule
    \end{tabular}
    \caption{\textbf{Results with worse reconstruction.} We simulate worse point matching case.}
    \label{tab:rob_rec}
\end{table}
\subsubsection{Robustness across various levels of 2D annotation quality.}
We discuss two kinds of worse 2D annotation qualities. The first is the influence of box numbers. We randomly drop 5\% 2D bounding boxes and add 5\% false positives. The second is for the 2D bounding box position. We add a maximum of 20\% position error to the 4 corner points of the 2D bounding box. The results are shown in Table~\ref{tab:rob_det}. Our method is relatively robust to the 2D box quality. Note that 2D object detection is a very stable and reliable technology. The 2D object detector is usually no worse than this simulation experiment. (For vehicles, $>$80 AP under 0.7 IoU threshold.)\par
\begin{table}[h]
    \centering
    \begin{tabular}{l|cccc}
    \toprule
        &3D AP$_\text{5}$&3D APH$_\text{5}$&LET APL$_\text{50}$&LET AP$_\text{50}$ \\ 
        \midrule
        GT box & 41.17 & 28.73 & 12.23 & 21.41 \\
        Add FP + FN & 36.96 & 26.24 & 9.34 & 16.51 \\
        Inaccurate box position & 31.34 & 21.88 & 8.45 & 14.81 \\
        \bottomrule
    \end{tabular}
        \caption{\textbf{Results using worse 2D annotations.}}
    \label{tab:rob_det}
\end{table}

\subsubsection{Necessity of Local Point Clustering (LPC).} LPC has two main roles: (1) provide semantic labels (class and ID) for reconstructed 3D points; (2) remove background points in the 2D box (2D box is not tight enough for the object boundary). If only using GPC to remove background points, the background points in frame A may be close to the foreground points in frame B (because there are more points, points are near to each other globally), and thus the clustering is not easy. Besides, without LPC, there will be more points to be clustered in GPC, which takes more memory and time. We conduct an additional ablation study (Table~\ref{tab:lpc}), i.e., we compare with using all points in 2D bounding box for GPC. The experiment also shows the effectiveness of our design of DoubleClustering.\par
\begin{table}[h]
    \centering
    \begin{tabular}{l|cccc}
    \toprule
        &3D AP$_\text{5}$&3D APH$_\text{5}$&LET APL$_\text{50}$&LET AP$_\text{50}$ \\ 
        \midrule
        GPC & 32.21 & 22.30 & 8.24 & 14.97 \\
        LPC+GPC & 41.17 & 28.73 & 12.23 & 21.41 \\ 
        \bottomrule
    \end{tabular}
    \caption{\textbf{Ablation study on DoubleClustering algorithm.}}
    \label{tab:lpc}
\end{table}

\subsubsection{Learning process details for $G_\theta$.}
\label{sec:gba_detail}
As mentioned in Section~\ref{sec:g_theta}, we use the length of pseudo 3D bounding boxes to determine whether the object is \emph{well-reconstructed}. This is a very simple yet effective metric, because we only need to find a good 3D box instead of its 3D surface in the object detection task. As mentioned in Section~\ref{sec:g_theta}, $G_\theta$ can improve box size estimation accuracy. Here, we report size prediction error, as the average absolute relative error of length, width and height in Table~\ref{tab:com_gba}. \par
\begin{figure}[h]
          \begin{center}
          	\includegraphics[width = 0.5\linewidth]{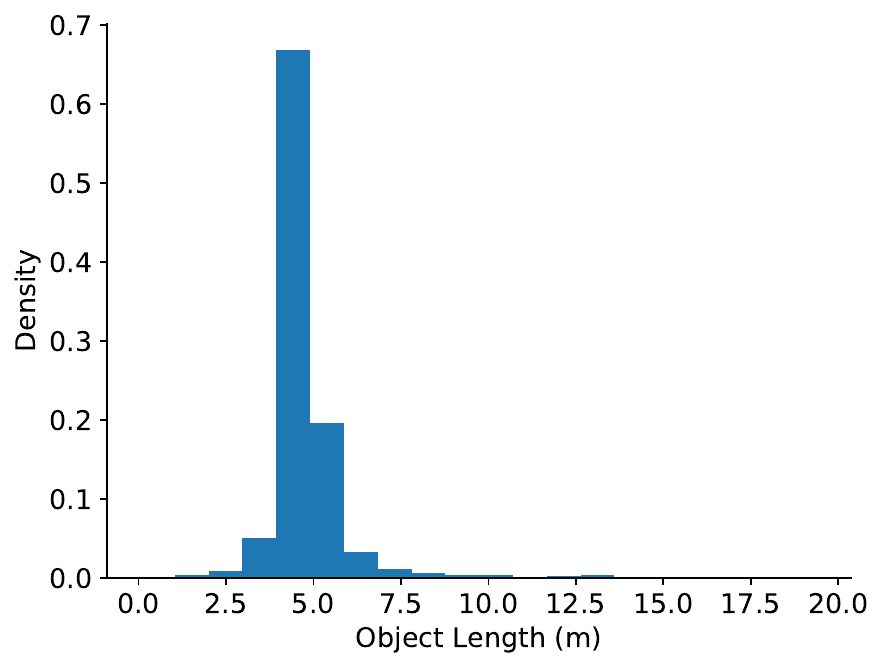}
          \end{center}
             \caption{\textbf{Statistical results of GT object length distribution.}}
             \label{fig:box_stat}
\end{figure} 
Besides, in training $G_\theta$, we do not simulate too large objects because we set a very loose length threshold. A vehicle whose length is too big ($>$10m) is very rare. According to the statistical results of GT (Fig.~\ref{fig:box_stat}), it shows that less than 1\% of objects have a length greater than 10m. We suppose that when the pseudo label is longer than 10m, it is likely that multiple objects have been incorrectly clustered into one cluster, resulting in a false positive. So we ignore this kind of cluster instead of learning a 3D box from it. As for the small pseudo box, we think it is a partially reconstructed object, and we can learn a full 3D box for it.\par
\begin{table}[h]
    \centering
    \begin{tabular}{l|cccc|c}
    \toprule
        &3D AP$_\text{5}$&3D APH$_\text{5}$ &LET APL$_\text{50}$&LET AP$_\text{50}$ &Avg. Abs. Rel.\\ 
        \midrule
        w/o $G_\theta$ & 28.40 & 11.34 & 5.02 & 8.62 & 0.076 \\ 
        w/ $G_\theta$ & 33.75 & 11.94 & 9.63 & 16.80 & 0.063 \\ 
        \bottomrule
    \end{tabular}
    \caption{\textbf{Ablation study of $G_\theta$.} We also report results under the size prediction error metric.}
    \label{tab:com_gba}
\end{table}

\subsubsection{Orientation error.}Our designs of orientation optimization (Sec.~\ref{sec:fit}) and orientation loss can help orientation esitmation. Besides APH metric, we also report more detailed orientation metrics. As for the orientation metrics, we report the average absolute relative error of orientation, defined as $\min (|\widetilde \theta-\theta|,2\pi-|\widetilde \theta-\theta|)/\pi$, where $\widetilde \theta$ and $\theta$ are the predicted heading and the ground truth heading. In Table~\ref{tab:ori_err}, we discuss the main factors to influence the orientation estimation, (1) minimizing the sum of distance (r$_y$ w/ d) and (2) orientation loss.\par
\begin{table}[h]
    \centering
    \begin{tabular}{l|cc|c}
    \toprule
        &3D AP$_\text{5}$&3D APH$_\text{5}$ &r$_y$ Abs. Rel.\\ 
        \midrule
        minAreaRectangle + MultiBin loss & 33.75 & 11.94 & 0.205 \\ 
        r$_y$ w/ d + MultiBin loss & 38.39 & 24.33 & 0.099 \\ 
        r$_y$ w/ d + Orientation loss & 41.17 & 28.73 & 0.072 \\ 
        \bottomrule
    \end{tabular}
    \caption{\textbf{Impact of orientation optimization and loss design for orientation estimation.}}
    \label{tab:ori_err}
\end{table}

\subsubsection{Moving object filtering in pseudo label generation.}The points on moving objects are mostly ignored in SfM. To further alleviate the effect of these points, we filter the object that has few points. We only keep the object cluster for more than $\theta$ = 100 points. These objects may be dynamic objects that are not reconstructed well. The influence of this operation is ablated in Table~\ref{tab:dyn_obj}.\par
\begin{table}[h]
    \centering
    \begin{tabular}{l|cccc}
    \toprule
        &3D AP$_\text{5}$&3D APH$_\text{5}$&LET APL$_\text{50}$&LET AP$_\text{50}$ \\ 
        \midrule
        ALL & 36.03 & 24.92 & 9.27 & 16.38 \\ 
        Filter $<$100 points & 41.17 & 28.73 & 12.23 & 21.41 \\
        \bottomrule
    \end{tabular}
    \caption{\textbf{Ablation study on few-point object filtering.}}
    \label{tab:dyn_obj}
\end{table}

\subsubsection{Pedestrian and cyclist categories.} Although we report the results of VEHICLE class in main paper, we would like to discuss the potential of \methodname{} for other categories as well. Other categories such as pedestrians and bicycles are primarily dynamic objects that would be affected by the moving object filtering operation in global scene reconstruction. \par
\begin{itemize}
    \item As for 3D pseudo-label generation, "the points on moving objects are mostly ignored". That means 3D pseudo labels are rarely generated for pedestrians and cyclists due to their movement. (However, when they are static, such as waiting for a red light, the pseudo label can still be generated.) This is the natural shortage of SfM. 
    \item Although lacking 3D pseudo labels, we can utilize the "generalization ability of depth from other objects" of the monocular 3D object detector. The other objects are mainly vehicles that are learned with 3D pseudo labels. We can generalize the depth of pedestrians/cyclists from the learned depth of vehicles. This generalization ability is because (1) Depth is the class-agnostic attribute of the object, and the network learns depth from the whole image. During the inference stage, the network can leverage the learned depth information of other vehicles in the same image to predict the depth of pedestrians and cyclists. (2) We train monocular 3D object detector with 2D assignment strategy. That means we will predict depth for all 2D objects, both vehicles and pedestrians/cyclists.
\end{itemize}
 We show the object-level depth and orientation estimation accuracy in Table~\ref{tab:ped_cyc}. As we can see, the average depth accuracy for pedestrians is no worse than vehicles. The training samples for the cyclist are too rare, and thus slightly affect the performance of the cyclist. \par
 \begin{table}[b]
    \centering
    \resizebox{\columnwidth}{!}{
    \begin{tabular}{l|ccccccc|c}
    \toprule
        Category&$\delta< 1.25\uparrow$  &$\delta< 1.25^2\uparrow$&$\delta < 1.25^3\uparrow$  &Abs Rel$\downarrow$ &Sq Rel$\downarrow$ &RMSE$\downarrow$ &RMSE log$\downarrow$ &r$_y$ Abs. Rel. \\ 
        \midrule
        Car & 0.993 & 0.995 & 0.996 & 0.093 & 0.558 & 3.367 & 0.198 &0.0805\\
        Pedestrian & 0.981 & 0.992 & 0.994 & 0.055 & 0.292 & 3.086 & 0.104 &0.0804 \\ 
        Cyclist & 0.848 & 0.950 & 0.962 & 0.120 & 0.804 & 4.491 & 0.168 &0.3737\\
        \bottomrule
    \end{tabular}}
    \caption{\textbf{Additional results on Pedestrian and Cyclist categories.} The middle columns show the object-level depth estimation results and the rightmost column shows the object-level orientation estimation results.}
    \label{tab:ped_cyc}
\end{table}

\section{Qualitative Results}
\label{sec:vis}
We show some qualitative results about 3D object detection and tracking (BA$^2$-Det), open-set 3D object detection, and 2D MOT with auxiliary 3D representation (\methodname{}). For more qualitative results and video demos, please refer to the project page: \url{https://ba2det.site}.
\subsection{3D Object Detection and MOT Results on WOD}
\begin{figure*}
          \begin{center}
          \resizebox{\linewidth}{!}{
          	\subfloat[BA-Det (10\% labeled videos).]{\includegraphics[width = 0.33\linewidth]{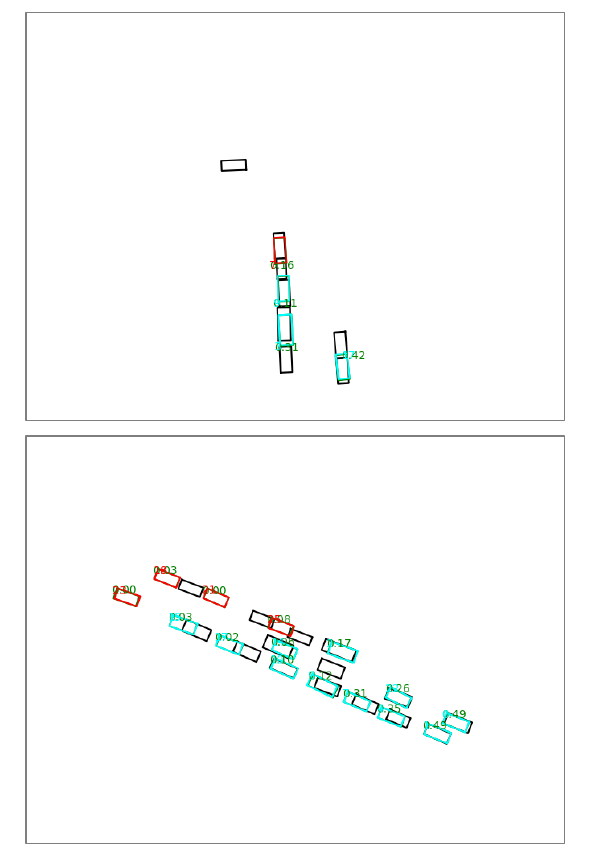}}
        	\hfill\ 
        	\subfloat[SfM + BA-Det (Baseline).]{\includegraphics[width = 0.33\linewidth]{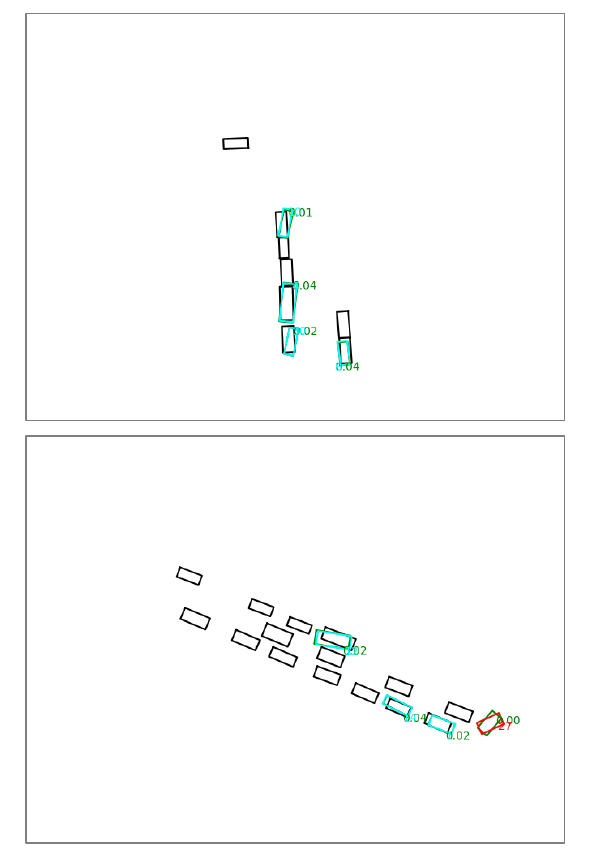}} 
         \hfill\ 
         \subfloat[BA$^2$-Det (Ours).]{\includegraphics[width = 0.33\linewidth]{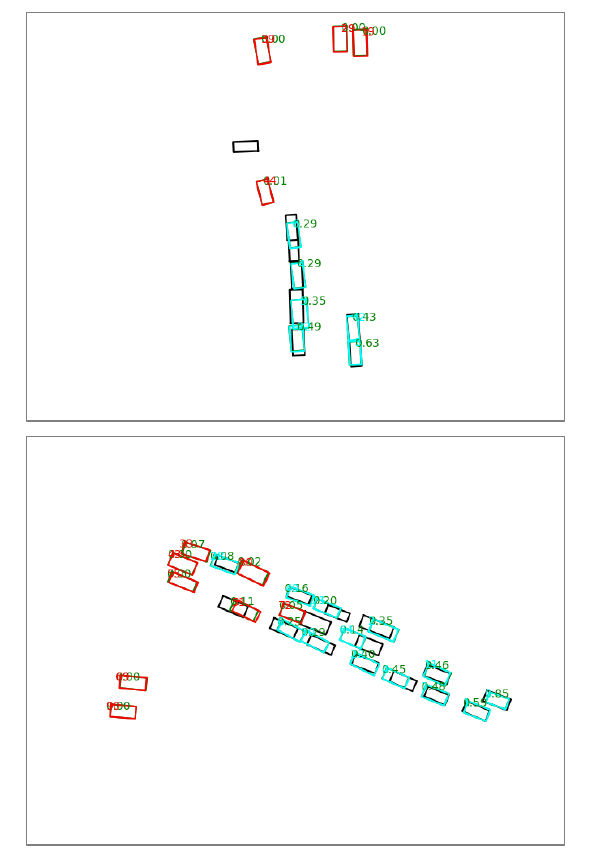}}
              }
          \end{center}
             \caption{\textbf{Qualitative results of 3D object detection and tracking shown in BEV.} \textbf{Black} boxes are the ground truth, \textbf{\textcolor{bluegreen}{cyan}} boxes are the tracking results with id, \textbf{\textcolor{darkgreen}{green}} boxes are the detection results with scores, \textbf{\textcolor{red}{red}} boxes are the false positives.}
             \label{fig:visdet}
\end{figure*}

In Fig.~\ref{fig:visdet}, we show the qualitative results of BA-Det (trained with 10\% labeled videos), the baseline method, and the proposed BA$^2$-Det. Our method can achieve comparable performance with fully supervised BA-Det, and even better in some near cases. Compared with the baseline, a very obvious phenomenon is that our recall can be much better than the baseline method, mainly due to the iterative self-retraining design. The illustrations also show a typical failure case of BA$^2$-Det that on a distance of about 75m, there are some false positives. This is because the 3D pseudo labels can be 0-200m and thus somewhat affects the training process. If the annotations include some farther objects, this problem may be alleviated.

\section{Limitations and Future Work}
\label{sec:limit}
We use 2D annotation to obtain pseudo 3D labels, and finally train a monocular 3D object detector with these pseudo labels. Similar to other monocular 3D detectors, we also encounter inherent issues such as inaccurate depth estimation and occlusion problems. The main contribution of this paper is proposing the use of weakly supervised methods to train a 3D detector. Compared to using 3D ground truth, our weakly supervised approach has the following limitations.
\begin{itemize}
    \item The limitation of reconstructing videos: Our method relies on reconstruction to obtain pseudo labels, so it is limited by the usage conditions of SfM. It is difficult to achieve good reconstruction results when there is no ego-motion or when the camera undergoes severe movement in the dataset. The method doesn't work well for purely static video datasets. However, less than 5\% of videos in the driving dataset do not meet the requirements, and we can use only pseudo labels with good reconstruction results to train the model. Therefore, this limitation can be greatly alleviated in driving scenarios. We also apply this method in more general scenes (Fig. 5), where most videos conform to the assumptions of this paper.
    \item The limitation of pseudo label quality: (1) Inaccurate orientation: Since many 3D clusters from the reconstructed scene are incomplete due to occlusion and self-occlusion, orientation estimation is more challenging than 3D location and 3D shape estimation. In section 3.2, we try our best to alleviate this problem. However, it still affects the performance and becomes an open problem. (2) Big cluster containing more than one object: When some objects are side-by-side, like in a parking lot, the clustering algorithm may output a big object cluster containing more than one object. We filter this kind of cluster by 3D shape threshold. In the future, we will improve the clustering algorithm to solve this problem. (3) Time-consuming iteratively self-retraining: Although iteratively self-retraining can improve performance a lot, it takes more than double the training time. We will try to decrease the time cost and keep the performance in the future work.
\end{itemize}
\section{Broader Impacts}
This paper presents work whose goal is to advance the field of Machine Learning. There are many potential societal consequences of our work, none which we feel must be specifically highlighted here.


\end{document}